\documentclass[runningheads]{llncs}

\usepackage{eccv}

\usepackage{eccvabbrv}

\usepackage{graphicx}
\usepackage{booktabs}

\usepackage[accsupp]{axessibility}  

\usepackage{hyperref}

\usepackage{orcidlink}

\usepackage{xcolor} 
\usepackage{listings}
\usepackage{graphicx}
\usepackage[hyphens]{url}
\usepackage{amsmath, amssymb}
\usepackage{booktabs}
\usepackage{algorithm}

\usepackage{algorithmicx}
\usepackage{algcompatible}
\usepackage{algpseudocode}

\usepackage{multirow}
\usepackage{hyperref}
\usepackage{geometry}
\usepackage{CJKutf8}
\usepackage{booktabs} 
\usepackage{lipsum}   
\usepackage{listings} 
\usepackage{xcolor}   
\usepackage[utf8]{inputenc}
\usepackage{wrapfig}
\usepackage{graphicx}   
\usepackage{array}      
\usepackage{multirow}
\usepackage[most]{tcolorbox}
\usepackage{colortbl}
\usepackage[T5]{fontenc}
\usepackage{selinput}
\SelectInputMappings{%
  adieresis={ä},
  eacute={é},
  Lcaron={Ľ},
  acircumflexdot={ậ},   
  acircumflexgrave={ầ}, 
  ecircumflexacute={ế}, 
}

\newtcolorbox{mybox}[2][]{
    colback=white,
    colframe=green!45,
    fonttitle=\bfseries,
    coltitle=black,
    fontupper=\scriptsize, 
    sharp corners,
    title=#2,
    #1
}

\lstdefinestyle{mypythonstyle}{
    language=Python,
    basicstyle=\ttfamily\small, 
    commentstyle=\color{gray},
    stringstyle=\color{red},
    showstringspaces=false,
    breaklines=true,              
    frame=tb,                     
    framerule=0.5pt,
    tabsize=4,
    captionpos=b                  
}

\algnewcommand\algorithmicinput{\textbf{Input:}}
\algnewcommand\Input{\item[\algorithmicinput]}
\algnewcommand\algorithmicoutput{\textbf{Output:}}
\algnewcommand\Output{\item[\algorithmicoutput]}

\newcommand*{\methodname}{LoGAN}

\definecolor{MyDarkBlue}{rgb}{0.02,0.02,0.6}

\definecolor{cellYellow}{HTML}{FFFFAA}
\definecolor{cellRed}{HTML}{F0A5A9}

\begin{document}

\title{LoGAN: Multilingual Font Localization with Generative Agents} 

\titlerunning{LoGAN}

\author{Zhuoning Yuan\inst{1}\orcidlink{0009-0009-9364-2486}, 
Ta-Ying Cheng\inst{1}\orcidlink{0000-0003-0717-6829}
,
Benjamin Klein\inst{1}\orcidlink{0009-0003-4069-7930}}

\authorrunning{Z. Yuan et al.}

\institute{\textsuperscript{1}Netflix 
}

\maketitle

\begin{abstract}
Localizing a font into new languages is a highly intricate task requiring precise design adaptation of glyphs, color/texture, and spacing/kerning, from source to target languages. Most existing methods focus on single glyph generation with limited capability in handling multilingual font rendering. In this work, we propose \methodname{}, a VLM-based agentic framework for few-shot multilingual font localization, which takes in a small number of individual glyphs from a font or letters from a logo and uses them to generate complete character sets in other languages. \methodname{} breaks down this task into multiple components: a glyph-level diffusion model, a style finetuning module, a spacing and kerning transfer algorithm, and a texture expansion model, with a VLM agent coordinator. \methodname{} achieves broad language coverage for font localization with various styles, including Chinese/Korean/Japanese (CJK). We evaluate our approach on both font and real-world logo datasets spanning more than 27 languages and compare it against both specialized font generation and state-of-the-art image editing models with strong text rendering capabilities (e.g., FLUX, Nano-Banana). Our approach yields higher glyph fidelity while maintaining better style, texture, and kerning consistency according to both quantitative and qualitative evaluations.

\end{abstract}    
\section{Introduction}
\label{sec:intro}

\begin{figure}[t] 
\centering 
\includegraphics[width=0.75\columnwidth]{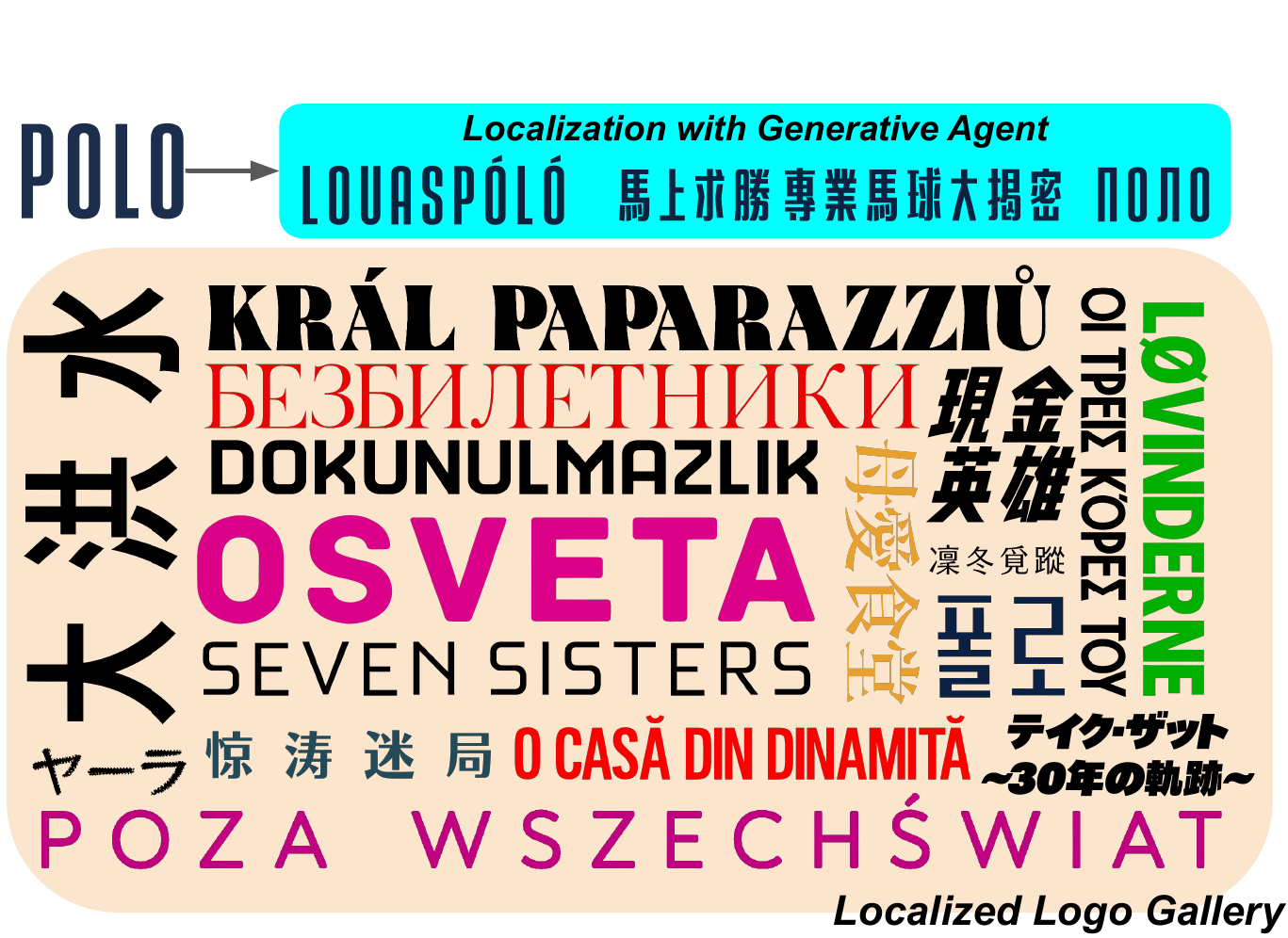}
\caption{\textbf{Overview.} We present \methodname{}, an agentic framework that localizes fonts into various languages. \methodname{} is able to generalize to various font styles, spacing/kernings, colors, and languages spanning across Latin, Greek, Cyrillic and CJK.} 
\label{fig:teaser} 
\end{figure}

Fonts are fundamental visual elements in graphic and logo design. However, most highly stylized fonts support a limited set of languages, predominantly offering only Latin character sets. Localizing these designs to new languages poses a significant challenge due to the immense creative freedom inherent in font design. Because artistic fonts frequently employ idiosyncratic geometric motifs, complex textures, and exaggerated stroke variations, transferring these unconstrained styles to entirely different character sets is highly non-trivial. Designers must navigate severe structural discrepancies—such as mapping simple Latin strokes to dense, complex CJK ideograms without sacrificing legibility—while preserving local stylistic motifs and re-engineering spatial dynamics like kerning. Consequently, in real-world workflows like movie logo localization, finding a matching multilingual font is a recurring bottleneck. When none is available, designers are forced into a difficult trade-off: either substitute a visually similar font and compromise stylistic coherence, or painstakingly synthesize the missing characters by hand. While movie logo creation involves graphical elements beyond basic fonts, our sample study of a 400-logo movie dataset (Section~\ref{sec:supp_vlm}) shows that 44\% of logos are created using fonts. Therefore, we focus specifically on the core challenge of font localization and use the movie logo localization task to evaluate our approach.

\begin{figure}[h] 
\centering
\includegraphics[width=\linewidth]{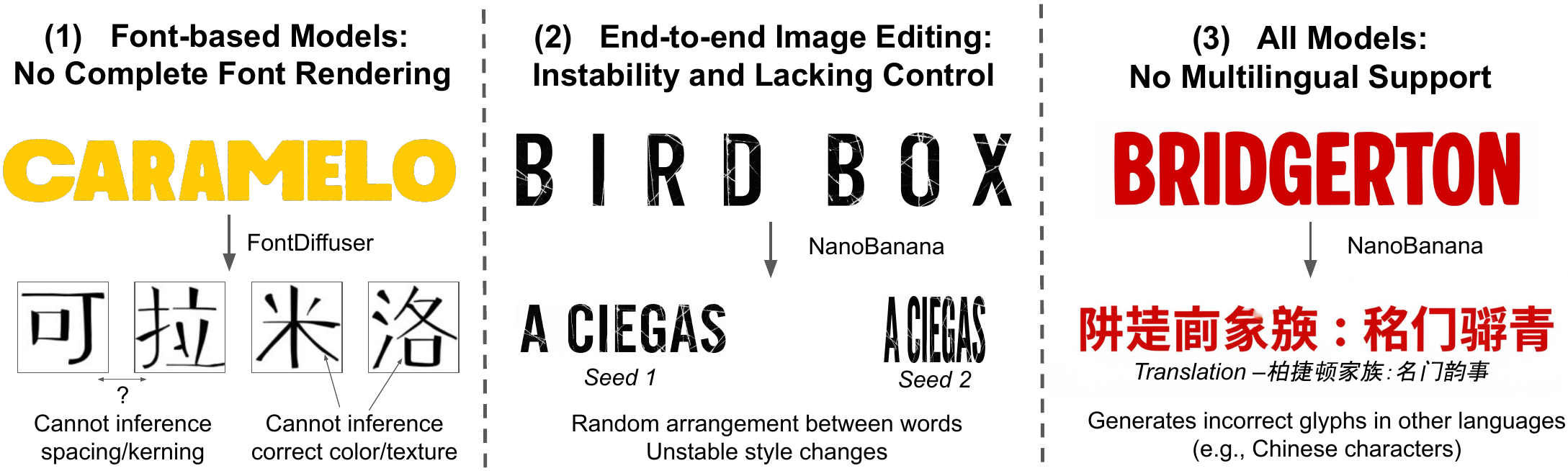}
\caption{\textbf{Challenges of current image generation models.} (1) Font-based models primarily focus on glyph generation and fails to render multiple characters with proper spacing/kerning and consistent color/textures. (2) End-to-end image editing models introduces randomness towards style changes and arrangements. (3) All current models lack a wide range of multilingual support.} 
\label{fig:teaser_2} 
\end{figure}

Recent advances in generative modeling—such as Nano-Banana, GPT-Image, FLUX, and Qwen-Image~\cite{google_gemini2_5flash,labs2025flux,wu2025qwen}—open up the possibility of applying general image diffusion models to tackle such challenges. A few pioneering works have explored using recent Text-to-Image diffusion models for creative design tasks, ranging from general image editing and style transfer~\cite{ye2023ip,google2025nanobanana,bfl2025flux}, poster-level generation~\cite{wang2025designdiffusion, chen2025posta,gao2025postermaker}, logo-level generation~\cite{tuo2024anytext2,chen2023textdiffuser,liu2024glyph, lan2025flux}, to font-level generation~\cite{li2023cross, li2024hfh, fu2024generate, huang2026vecglypher}. Although these methods show promising results, they are not readily adaptable to the challenges posed by font localization tasks due to the following: \textit{(1) Existing font generation models primarily focus on glyph generation}. Most prior work (e.g., FontDiffuser, VecGlypher) focuses on generating single glyph images or SVGs and \textbf{lacks the ability to render a complete font} (e.g., generating multiple characters with proper spacing/kerning and consistent color/texture, as shown in Figure \ref{fig:teaser_2} left column). This limitation makes these methods unsuitable for end-to-end practical use and more appropriate as intermediate steps. \textit{(2) Many end-to-end models can render complete text but lack controllability and stability}. For example, approaches such as FLUX and Nano-Banana often struggle with long-text rendering in font generation settings. When the target layout is complex, they may rescale letters improperly and produce incorrect arrangements (Figure \ref{fig:teaser_2} middle column). \textit{(3) Many existing approaches lack multilingual support}. Most recent image generation models, including font generation models, are primarily trained on a single language or only a few languages (e.g., English/Latin) and often fail to produce high-quality, consistent results for other scripts, such as Cyrillic, Greek, and CJK (Figure \ref{fig:teaser_2} right column). Additionally, they often struggle with cross-language localization tasks due to limited training data and/or insufficient cross-lingual modeling.

To address these challenges, we introduce a Multilingual Font \textbf{Lo}calization with \textbf{G}enerative \textbf{A}ge\textbf{n}t (\textbf{\methodname{}}). The key ideas behind LoGAN are twofold: \textbf{First}, while end-to-end models are impressive in one-shot generation, they struggle with output precision and controllability for font rendering tasks. Multiple design elements, such as glyphs, spacing/kerning, color/texture, and layout, are coupled in a one-pass generation, which often leads to hallucinations of certain design attributes and requires re-prompting and regenerating until the results are correct. Inspired by the strong visual understanding capabilities of recent SOTA VLMs (e.g., \texttt{GPT-5}, \texttt{Gemini-3-Pro}~\cite{ openai_gpt5_2025, google_gemini_2026}), we propose a divide-and-conquer approach by decomposing the complex font localization task into basic design elements. We further develop \textit{expert models} to fulfill each of these steps—such as using a VLM to detect color, spacing, and layout, using an image generation model to generate glyphs, and using a VLM agent to coordinate the complete font rendering process. \textbf{Second}, we focus explicitly on glyph-level generation followed by a spacing/kerning transfer module instead of word- or sentence-level text generation, which usually carries the risk of hallucinating certain prompts in underrepresented languages or new words created by the internet that do not frequently occur in the training data, as well as the risk of characters being rescaled during long-text rendering. We develop a strong multilingual glyph-level generation model that supports 27 languages for both same-language and cross-language generation, especially for CJK (see examples in Figure~\ref{fig:teaser}), as part of the proposed font localization pipeline.

To the best of our knowledge, this is the first work to propose an agent-based end-to-end pipeline for multilingual font localization and rendering. The design overview of LoGAN is shown in Figure~\ref{fig:LoGAN_design_overview}. Our main contributions are summarized as follows: We develop a novel framework, \methodname{}, that utilizes a VLM as a design agent to coordinate complex font localization and rendering tasks through three stages: design decomposition, design transfer, and design generation. We further develop a multilingual glyph generation module—including glyph generation, arrangement, and texturing models—for high-quality vector font rendering. We conduct a comprehensive evaluation to validate \methodname{}, spanning Latin, Cyrillic, Greek, and CJK languages on font and real movie logo datasets. We present both qualitative and quantitative results and compare \methodname{} against specialized font-generation methods and state-of-the-art image editing/generation models, achieving the best performance across all metrics.

\section{Related Work}
\label{sec:related_work}
Generative models have advanced substantially in the quality of image generation~\cite{rombach2022ldm, podell2024sdxl, liu2023rectifiedflow, esser2024scaling, ramesh2021dalle} and editing~\cite{brooks2023instructpix2pix,bfl2025flux, wu2025qwen, google2025nanobanana}. 

\textbf{Image Diffusion for Font Generation/Editing}. With these improvements, a plethora of works has started to focus on the controllability of diffusion models. These controls span structural guidance~\cite{zhang2023controlnet, bhat2023loosecontrol, wang2024instancediffusion}, style transfer~\cite{ye2023ip, wang2024instantstyle, chung2024styleid}, and visual characteristics/textural transfer~\cite{zhou2025attentiondistill, cheng2024zest, cheng2025marble}. However, these models fall short in the task of logo localization, where the target style must be precisely transferred while being disentangled from the original content. For example, direct style transfer or prompt-based image editing often entangles glyph content with stylistic attributes, bringing unwanted stroke structures and character-specific features into the newly localized logo. Several works aimed specifically at poster design~\cite{wang2025designdiffusion, chen2025posta, gao2025postermaker}, text rendering~\cite{tuo2024anytext2,lan2025flux,liu2024glyph} or font generation~\cite{fu2024msdfont,liu2024qtfont, yang2024fontdiffuser}. Nevertheless, their style transfers tend to be comparatively coarse and are less capable in generalizing across scripts with drastically different typographic and morphological properties, e.g., Latin-based scripts to Korean.

\textbf{VLM for Image Editing}.
Guiding diffusion models for optimal image generation remains an ongoing challenge. With the rise of LLMs and VLMs, many works leverage language model feedback to enhance prompts for generation~\cite{fu2024mgie,  huang2024smartedit, lian2024lmd}. These are all different approaches aiming to improve the prompt for better image generation.~\cite{yeh2025xplanner} This method decomposes an abstract prompt into sequences of simpler editing instructions for diffusion models.~\cite{wu2024sld} Another method improves the generated output based on the VLM feedback loop.~\cite{wu2025qwen} This approach feeds the prompt into a VLM and uses the hidden features as inputs into the diffusion model for greater expressibility. There are also approaches that treat VLM as an agent to call a library of model tools to iteratively refine a generated image~\cite{wang2024genartist}. 

In this work, we leverage VLM as a design agent that decomposes complex font localization into simpler tasks and coordinates specialized models to produce detailed, high-quality logos.
    
\section{\methodname{}}
\label{sec:approach}
\begin{figure}[t] 
    \centering 
    \includegraphics[width=\linewidth]{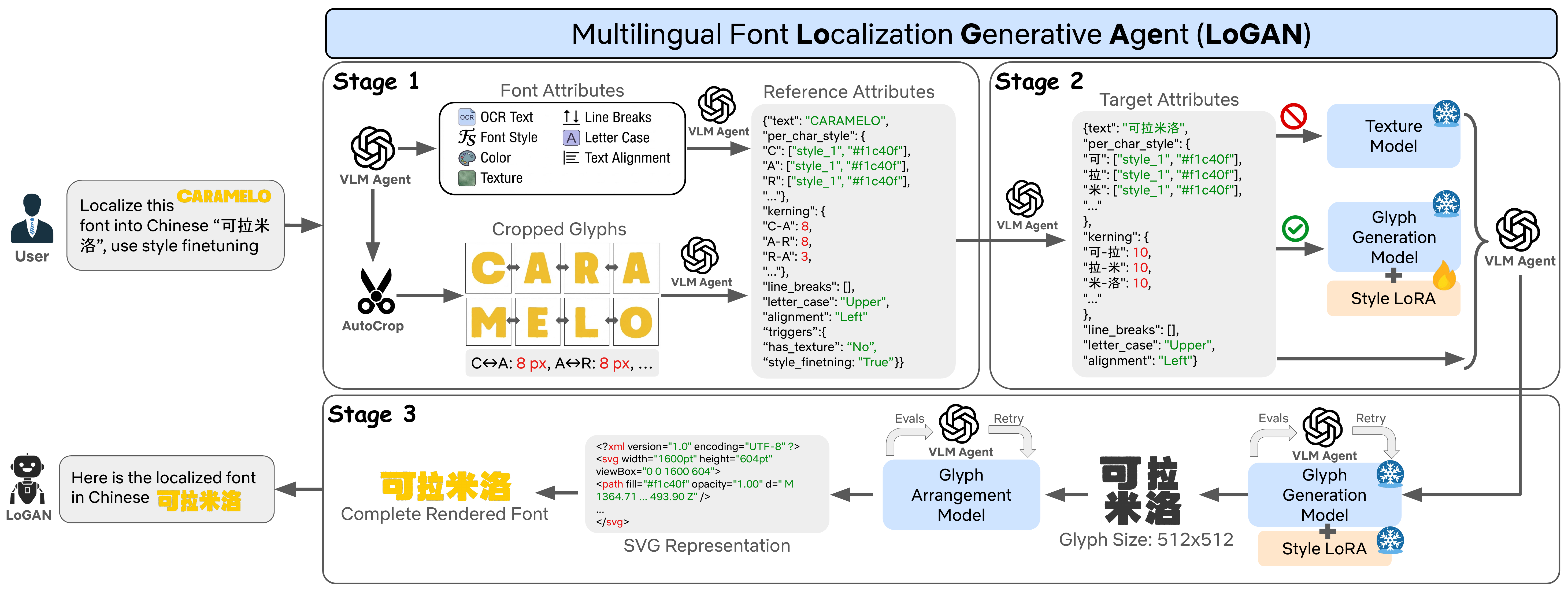} 
    \caption{\textbf{Method Overview.} \methodname{} takes a reference image and a target translation generates translated font with three stages: (1) Design Extraction, (2) Design Transfer, and (3) Design Generation. Style LoRA is trained during (2) and then frozen for inference during (3).} 
    \label{fig:LoGAN_design_overview} 
\end{figure}

We present the overall design and workflow of \methodname{}, as shown in Figure~\ref{fig:LoGAN_design_overview}, and introduce its key components in the subsequent sections. \methodname{} takes a base prompt of \texttt{Localize this image to [Translation]} and a reference font image, then executes the following actions: \textit{design extraction}, \textit{design transfer}, and \textit{design generation}, and finally returns a localized rendered font image that matches the style of the reference and target translation. For each stage, the details are as follows: 

\begin{itemize} 

\item 
\noindent\textbf{Stage 1: Design Extraction}. \methodname{} first decomposes users' input into: reference image, translation, and style finetuning trigger. Then, it calls VLM-Agent to analyze and extract design attributes from the reference image, including OCR text, font style, color/texture, line breaks, letter case, and text alignment, where texture and font style are in abstract form, e.g., \textit{style1}, \textit{texture1}. Next, it calls the AutoCrop module to crop the reference image into glyphs and measures the spacing between adjacent cropped glyph pairs. Finally, \methodname{} summarizes the extracted data as a \textit{character-attribute} mapping, key attributes, a list of cropped glyphs, and the trigger for style finetuning. 

\item 
\noindent\textbf{Stage 2: Design Transfer}. \methodname{} processes the reference attributes and triggers extracted in Stage~1. It then calls the VLM-Agent to infer a new \textit{character--attribute} mapping for the target translation and, optionally, triggers the Style-Finetuning module (based on the user's request) to fine-tune the pre-trained glyph generation model with LoRA~\cite{hu2022lora}, improving style transfer quality. Next, \methodname{} determines whether to call the Texture-Expansion module based on the texture detected in Stage 1. 

\item
\noindent\textbf{Stage 3: Design Generation}. With all required inputs, \methodname{} triggers the Glyph-Generation model, optionally with LoRA, using the translation to generate the target glyphs at a 512x512 resolution. It then calls the Glyph-Arrangement model to arrange the generated glyphs into a fully rendered font. Following this, \methodname{} calls VLM-Agent to inspect the outputs (e.g., glyphs, line breaks, colors), and triggers a retry if errors are detected. Finally, Glyph-Arrangement model vectorizes the generated font. If generated texture was detected, \methodname{} will overlay this texture map onto the final vectorized font.
\end{itemize}

\subsection{Key Components}
\noindent\textbf{VLM Agent.} This is the coordinator of LoGAN with three different roles: (1) \textit{Interpreter}: decompose the user request, extract design attributes from the reference image (e.g., OCR text, layout, style, color/texture, spacing/kerning, etc.), and assist with glyph cropping. (2) \textit{Planner}: localize the reference attributes to target languages, decide whether to trigger optional modules (e.g., style LoRA, Texture-Expansion) and trigger retries if evaluation fails. (3) \textit{Judge}: verify the correctness of the generated glyphs and the overall quality of font rendering (e.g., kerning/layout, color/texture consistency). In our implementation, we use \texttt{GPT-5.2}~\cite{openai_gpt5_2025} as the default VLM-Agent and validate that open‑source Qwen3-VL-32B-Instruct~\cite{Qwen3-VL} achieves comparable performance. \\

\noindent\textbf{Glyph Generation.} \label{sec:texture} Most  existing glyph/font generation approaches focus on only a few languages~\cite{yang2024fontdiffuser,huang2026vecglypher}. To address this, we develop a glyph generation model spanning 27 languages based on MMDiT (SD3.5-medium)~\cite{stabilityai_stable_diffusion_35_medium}. Specifically, we modify the joint attention in each block to add a third modality that encodes reference glyphs: we apply modality-specific projections (and RMSNorm), concatenate all modalities’ Q/K/V, followed by a single unified scaled dot-product attention~\cite{li2025omniflow}. For the other dependent components, we adopt siglip2-so400m-patch16-256~\cite{tschannen2025siglip} to encode reference glyphs, concatenate $k$ glyph embeddings into $(N, k*256, 1152)$, and pass them through Perceiver Sampler~\cite{alayrac2022flamingo} to obtain a fixed-length embedding $(N, 256, 1152)$. We utilize ByT5-Large~\cite{xue2021byt5} for glyph-level prompt encoding and a 16-channel VAE from AuraDiffusion~\cite{AuraDiffusion_16ch-vae}. We finally train a 2B-parameter glyph generation model from scratch. Training details are included in Section~\ref{sec:glyph_pretrainined}. We also develop a Style-Finetuning module with LoRA finetuning on reference glyphs to ensure the generated glyphs preserve the exact character identity when seen in the reference (e.g., \textbf{Zootopi}a[EN] $\rightarrow$ \textbf{Zoot}r\textbf{opo}l\textbf{i}s[TR]), as most zero-shot approaches often introduce subtle pixel shifts that can limit practical usability. Additionally, we find this fine-tuning can improve overall style transfer quality (Table~\ref{tab:results_category}). \\

\noindent\textbf{Glyph Arrangement.} \label{sec:arrangement} Given the generated glyphs, the reference/target design attributes (e.g., line breaks, spacing pairs), \methodname{} arranges the generated glyphs into a fully rendered font that matches the reference through four sub-tasks: (1) spacing/kerning transfer, (2) line break and alignment transfer, (3) glyph vectorization, and (4) color/texture transfer.
\begin{itemize}
\item \textbf{Spacing and Kerning Transfer.} Kerning and spacing refer to the gap between two adjacent letter pairs. Different character pairs require different spacing to achieve a similar perceived visual gap (e.g., ``WA'' is typically much tighter than ``HO''). \methodname{} predicts spacing for target character pairs based on the spacing of reference character pairs extracted by AutoCrop.  The key challenge is to reuse the same character-pair spacings when they are present in the reference and to infer unseen pairs using typographic rules. Detailed steps are as follows:

\begin{table}[t]
\centering
\small 
\caption{Abstract Shapes for Kerning Heuristics. }
\begin{tabular}{|l|p{8cm}|}
\hline
\textbf{Shape (Symbol)} & \textbf{Description} \\
\hline
Straight ($\mathbf{|}$) & Vertical or implied vertical boundaries that restrict close kerning (e.g., both sides of ``H'', right side of ``E'', left side of ``B'', etc. All CJK letters are also assumed to fall within this shape). \\
\hline
Circular ($\mathbf{\circ}$) & Curved boundaries that naturally accommodate tighter kerning (e.g., both sides of ``O'' or the left side of ``C''). \\
\hline
$\nearrow$ Diagonal  ($\mathbf{/}$) & Ascending diagonal strokes from bottom-left to top-right (e.g., the left side of ``A'' or the right side of ``V''. Right side of ``T'' also belongs to this shape as you can draw an implicit $\nearrow$ for tighter kerning). \\
\hline
$\searrow$ Diagonal  ($\mathbf{\backslash}$) & Descending diagonal strokes from top-left to bottom-right (e.g., the right side of ``A'' or the left side of ``V''). Certain lowercase letters, such as ``e'' or ``w'', are also treated as diagonal because their open contours provide spatial gaps for tighter kerning. \\
\hline
\end{tabular}
\label{tab:kerning_shapes}
\end{table}

\begin{itemize}
\item We first categorize the left and right contour of each character (across languages) into four abstract shapes (Table~\ref{tab:kerning_shapes}): Straight ($\mathbf{|}$), Circular ($\mathbf{\circ}$), $\nearrow$ Diagonal ($\mathbf{/}$), and $\searrow$ Diagonal ($\mathbf{\backslash}$). This abstraction maps character pairs to high-level shape combinations. For example, both ``HO'' and ``EC'' correspond to $\mathbf{|}\mathbf{\circ}$. Based on a small study with typographic experts, we adopt the following heuristic ordering:
 
\begin{equation}
\begin{split}
S \left(\mathbf{| |}\right) &> S \left(\mathbf{| /}\right) = S \left(\mathbf{| \backslash}\right) > S \left(\mathbf{| \circ}\right) = S \left(\mathbf{\circ \circ}\right) > S \left(\mathbf{\backslash /}\right) \\
&> S \left(\mathbf{\circ \backslash}\right) = S \left(\mathbf{\circ /}\right) > S \left(\mathbf{\backslash \backslash}\right) = S \left(\mathbf{/ /}\right),
\end{split}
\label{eq:spacing_heuristics}
\end{equation}

 where $S(\cdot)$ denotes the spacing/kerning for a boundary-shape pair (right side of the preceding character + left side of the succeeding character).

\item For each character pair in the reference, we extract watertight, character-level bounding boxes using AutoCrop (Section~\ref{sec:auto_crop}). We then measure the pixel gap between bounding boxes for each character pair and store these values in a dictionary. If multiple character pairs in the reference share the same abstract shape pair, we randomly select one measured value and apply it to the corresponding target shape pair. During inference, we reuse the cached value when the same character pair appears in the reference; otherwise, we infer the spacing using Eq.~\ref{eq:spacing_heuristics}. Finally, we refine the result by enforcing a minimum pixel gap between adjacent glyphs, bounded by the average spacing across the localized font.

\end{itemize}
\item \textbf{Line Break and Alignment Transfer.} We apply line breaks based on positions -- the character indices at which to insert line breaks, e.g., “Hello <br> World”, where the break index is 5 -- and text alignment (left|center|right) from target design attributes inferred by VLM-Agent.

\item \textbf{Glyph Vectorization.} We convert generated glyphs to SVG to make them editable for future use. Following~\cite{li2024hfh}, we use VectorMagic~\cite{cedarlake_vectormagic} and also explore an open-source alternative, Potrace~\cite{selinger_potrace_2019}, which has comparable performance. We compare different PNG-to-SVG approaches (see more examples in the Appendix) and find that directly generating accurate SVGs remains challenging~\cite{rodriguez2024starvector}.

\item \textbf{Color and Texture Transfer.} To transfer monochrome colors, we directly apply the color code from the reference to the final vectorized font. For complex textures, we feed the image containing the textured text into the Texture-Expansion model to recover the texture map from the given reference font (Figure~\ref{fig:texture}) and overlay it onto the final vectorized font.
\end{itemize}

\begin{wrapfigure}{r}{0.5\columnwidth}
  \centering
  \vspace{-0.3in}
  \includegraphics[width=0.5\columnwidth]{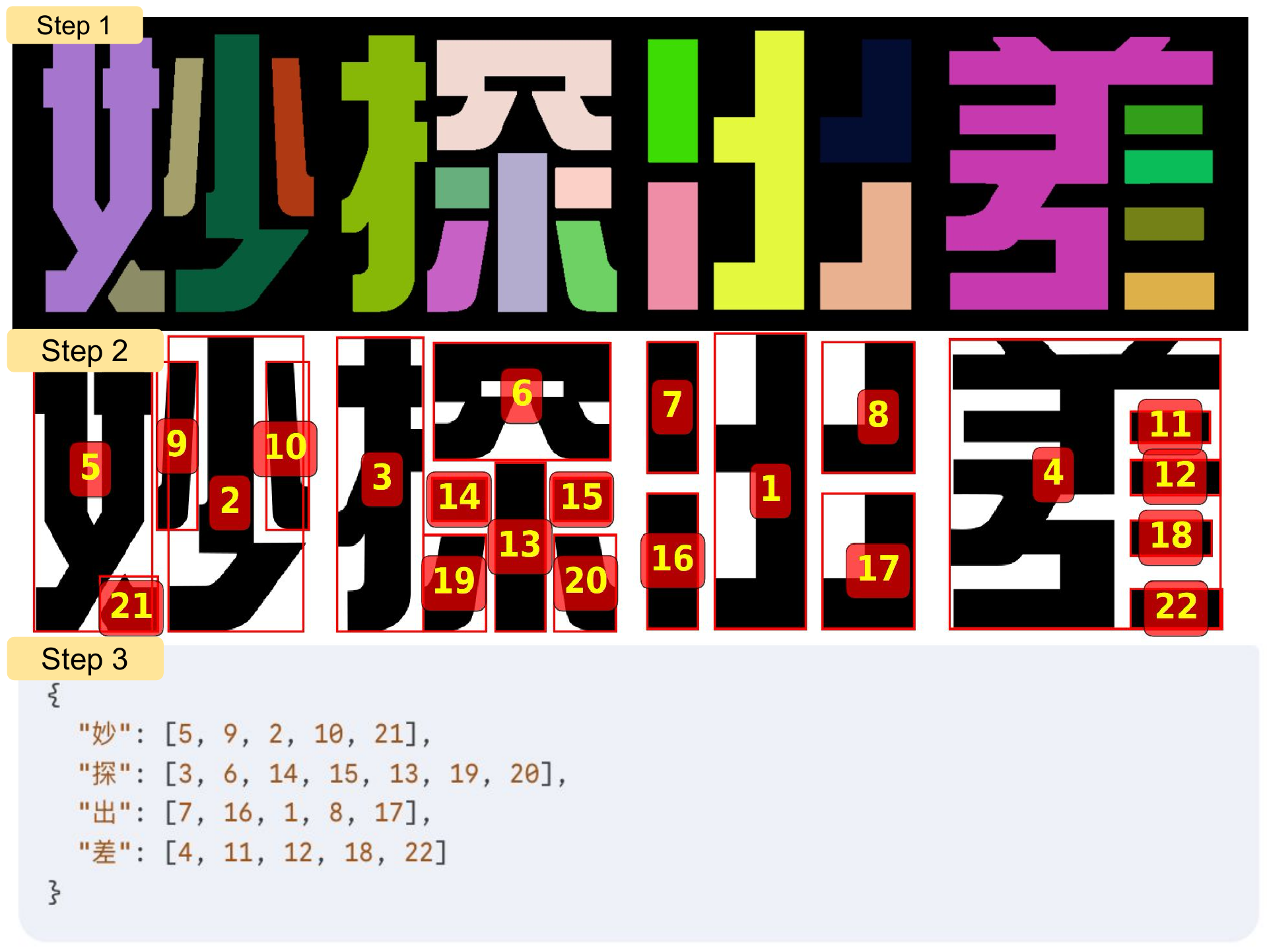}
  \caption{Illustration of character-level cropping for Chinese characters ``Beverly Hills Cop".}
  \label{fig:auto_crop}
  \vspace{-0.3in} 
\end{wrapfigure}

\noindent\textbf{Glyph Cropping.} \label{sec:auto_crop}
One key challenge in both glyph generation (constructing model inputs) and glyph arrangement (measuring inter-character spacing) is to reliably crop a complete rendered font into individual glyphs. Most prior text-detection methods operate at the word or line level~\cite{cui2025paddleocr30technicalreport, wei2025deepseek, li2025monkeyocrdocumentparsingstructurerecognitionrelation}. Although CRAFT~\cite{baek2019character} performs character-level detection, it is trained on SynthText and ICDAR~\cite{gupta2016synthetic} which can lead to the domain shift for our setting. We therefore propose a training-free method for multilingual character-level detection and cropping named AutoCrop. Our method is robust to characters composed of multiple disconnected components (e.g., CJK characters or Latin characters with diacritics)~\cite{cui2025paddleocr30technicalreport}. AutoCrop consists of four steps: (1) we apply connectedComponentsWithStats~\cite{wu2009optimizing} to extract connected components and their masks; (2) we render an index label on each component; (3) we use VLM (OCR) to group component indices by character identity and then merge grouped components to produce a refined bounding box per character; and (4) we sort refined bounding boxes in reading order (top-to-bottom, left-to-right) to align with the OCR transcript. Figure~\ref{fig:auto_crop} illustrates the overall pipeline.


\subsection{Training Strategy}
\label{sec:training_strategy}

\begin{wrapfigure}{r}{0.52\columnwidth}
  \vspace{-0.3in} 
  \centering
  \includegraphics[width=0.50\columnwidth]{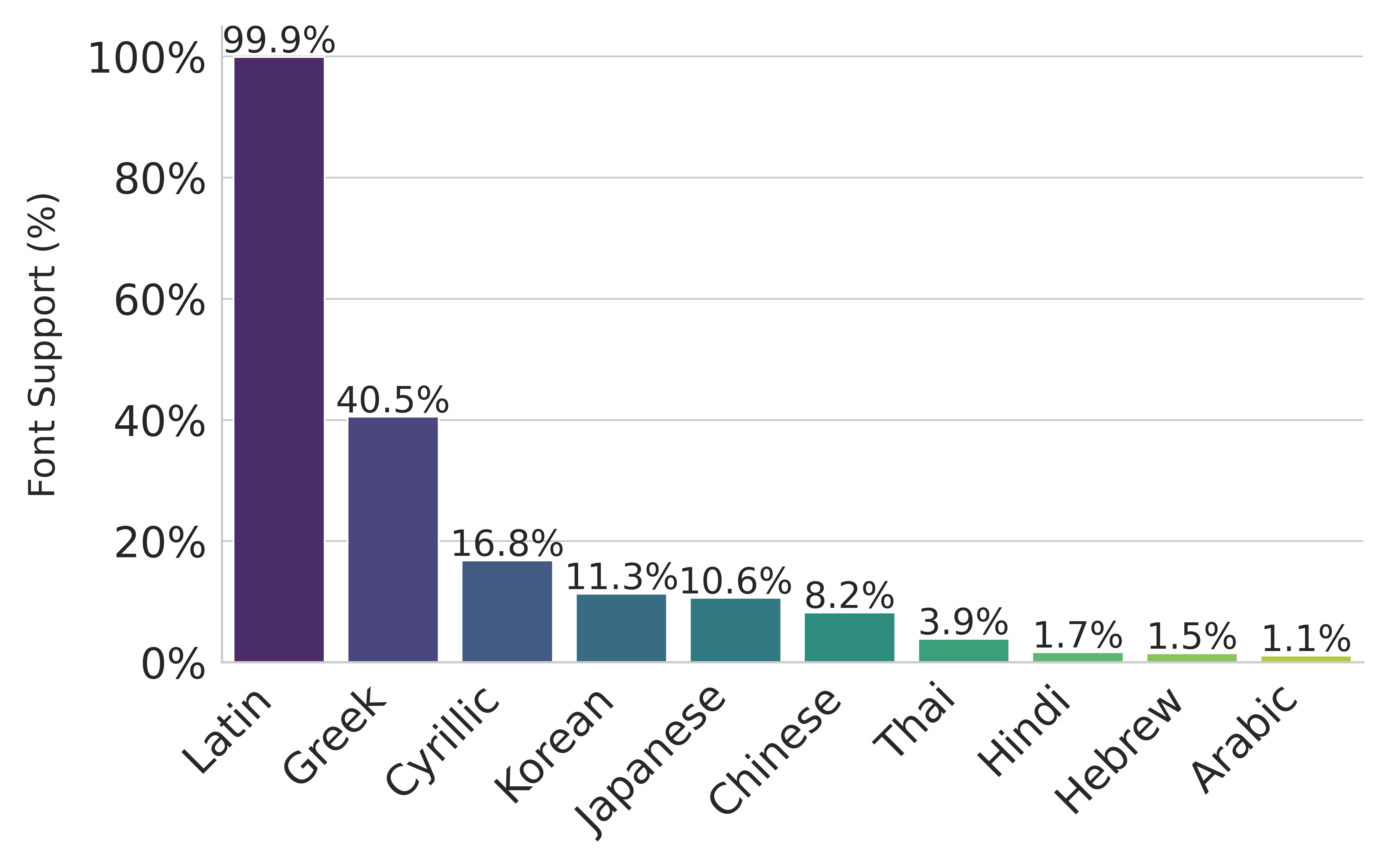}
  \caption{Statistics of language distribution on our internal font dataset.}
  \label{fig:font_language_distribution}
  \vspace{-0.3in}
\end{wrapfigure}

\noindent\textbf{Cross-Lingual Sampling}. Many existing generation models~\cite{bfl2025flux, google2025nanobanana, yang2024fontdiffuser, huang2026vecglypher} are specialized in a limited number of languages due to the scarcity of  multilingual training data and limited cross-lingual modeling. To address these challenges, we first analyzed an internal font dataset of 15,000 popular fonts and found a drastically long-tailed distribution that favors Latin languages over others (Figure~\ref{fig:font_language_distribution}). We then propose a cross-lingual sampling algorithm, presented in Algorithm~\ref{alg:lang-aware} and~\ref{alg:case-sensitive}. The core idea is to mitigate the long-tail bias by explicitly balancing same-language and cross-language sampling pairs during training. For Latin scripts, we further introduce case- and diacritic-aware sampling to improve training diversity, e.g., mixing letter cases (hELlO $\rightarrow$ A) and mapping diacritic variants to their base forms (AÀÁÂÄÅ $\rightarrow$ A, ùúûüũū $\rightarrow$ u). Finally, we include some duplicated pairs (AAAAAA $\rightarrow$ A) to further increase training diversity. \\

\noindent\textbf{Glyph Generation Training.} \label{sec:glyph_pretrainined} We started with pre-training on 256$\times$256 image resolution using Fully Sharded Data Parallel (FSDP) distributed across 8 A100 (80GB) GPUs with bfloat16 mixed precision and a global batch size of 1024. We freeze all encoders (image encoder, VAE) and train only the main transformer backbone. We follow the training settings from Stable Diffusion 3~\cite{esser2024scaling}. We utilized AdamW optimizer with $\epsilon=10^{-15}$. The learning rate was fixed at $1.0 \times 10^{-4}$ using a constant scheduler, following a linear warmup  of 1,000 steps. We applied gradient clipping with a maximum norm of $1.0$. After training for 450k iterations, we switched to 512$\times$512 image resolution for training another 450k iterations using the same learning parameters, with the batch size reduced to 640. Additionally, we used three types of glyph augmentation: zooming in or out, widening or narrowing either horizontally or vertically, and applying italic styles following~\cite{li2024hfh}. The total training takes about 15 days to complete. \\

\noindent\textbf{Style Finetuning.} Starting from the pretrained 2B glyph generation model, we train a LoRA for Style-Finetuning. We first use AutoCrop to crop the reference font into square images for each glyph. Since the reference font can be of arbitrary sizes, we resize all cropped glyphs to 512×512. We run the LoRA fine-tuning on a single A100 (40GB). We use PEFT~\cite{peft} to implement LoRA with rank 64 and alpha 64 with about 159.5M trainable parameters. The injected LoRA layers are included in the Appendix. We use the same training configurations as in pretraining, except with a smaller batch size (up to 8), depending on the number of available cropped glyphs in the reference. We train the LoRA for 800 steps, and the average training time is about 30--40 minutes. \\

\noindent\textbf{Texture Expansion Model.} We train a LoRA~\cite{hu2022lora} on the FLUX.1-Fill-dev model~\cite{blackforestlabs_flux1_fill_dev} to recover the full texture given a reference font with an overlaid texture. We use PEFT to implement the LoRA fine-tuning. The trainable parameters are about 44.8M. Given the relatively small dataset size of PBR materials~\cite{FreePBR}, we apply heavy augmentations to the materials, including zooming, changing brightness, color inversion, etc., to expand texture diversity. We also randomize font sizes and dynamically sample characters on-the-fly to further vary stroke widths and spatial layouts. We used a global batch size of 2 and a learning rate of 1e-4 for training 20,000 steps distributed across 2 A100 (40GB) GPUs using bfloat16 mixed precision. We use an image resolution of 1536x1536 for training. The total training takes about 12 hrs to complete.

\begin{table}[t]
\centering
\captionsetup{type=algorithm}
\begin{minipage}[t]{0.49\linewidth}
\begin{algorithm}[H]
\scriptsize  
\caption{Language-Aware Char Sampling}
\label{alg:lang-aware}
\begin{algorithmic}[1]
\State \textbf{Input:} Character $c$, Font $f$, Sample size $k$
\State \textbf{Output:} Set of $k$ sampled characters
\State \textbf{Notation for Algorithms~1 and~2:}
$L(c,f)$ returns the language group of character $c$ in font $f$;
$F(l,f)$ returns the set of all characters in language group $l$ for font $f$;
$\mathcal{L}=\{\text{Latin, Greek, Cyrillic, Chinese, Japanese, Korean}\}$.
\Call{IsUpper}{c} and \Call{IsLower}{c} return \texttt{true} if character $c$ is uppercase or lowercase, respectively.
\Call{Normalize}{c} removes diacritics from character $c$.
\Call{Denormalize}{c} returns all diacritic variants of character $c$.

\Function{SampleCharacters}{$c, f, k$}
    \State $p_1, p_2 \gets \Call{Random()}{}, \Call{Random()}{}$
    \State $l_1 \gets L(c, f)$
    \If{$p_1 < 0.05$}
        \State \Return $[c]\times k$
    \EndIf
    \If{$p_1 < 0.5$}
        \State $\mathcal{C} \gets F(l_1, f)$ \Comment{Same-language sampling}
    \Else
        \State $l_2 \gets \Call{RandomSample}{\mathcal{L} \setminus \{l_1\}, 1}$
        \State $\mathcal{C} \gets F(l_2, f)$ \Comment{Cross-language sampling}
    \EndIf
    \State \Return \Call{CaseSensitiveSample}{$c, k, \mathcal{C}, p_2$}
\EndFunction
\end{algorithmic}
\end{algorithm}
\end{minipage}\hfill
\begin{minipage}[t]{0.49\linewidth}
\begin{algorithm}[H]
\scriptsize  
\caption{Case-Sensitive Char Sampling}
\label{alg:case-sensitive}
\begin{algorithmic}[1]
\State \textbf{Input:} Character $c$, Sample size $k$, Candidate set $\mathcal{C}$, Probability $p_2$
\State \textbf{Output:} Set of $k$ sampled characters
\Function{CaseSensitiveSample}{$c, k, \mathcal{C}, p_2$}
    \If{$p_2 < 0.25$}
        \If{\Call{IsUpper}{$c$}}
            \State $\mathcal{C} \gets \{x \in \mathcal{C} \mid \Call{IsLower}{x}\}$
        \ElsIf{\Call{IsLower}{$c$}}
            \State $\mathcal{C} \gets \{x \in \mathcal{C} \mid \Call{IsUpper}{x}\}$
        \EndIf
    \ElsIf{$0.25 \le p_2 < 0.5$}
        \State $\mathcal{C} \gets \{c + x \mid x \in \mathcal{C}\}$ \Comment{Must include $c$}
    \ElsIf{$0.5 \le p_2 < 0.75$}
        \If{\Call{HasDiacritic}{$c$}}
            \State $c_{\text{base}} \gets \Call{Normalize}{c}$
            \State $\mathcal{C} \gets \{x + c_{\text{base}} \mid x \in \mathcal{C}\}$
        \Else
            \State $\mathcal{C} \gets \{x + c \mid x \in \mathcal{C} \land \Call{Denormalize}{x}\}$
        \EndIf
    \EndIf
    \If{$|\mathcal{C}| < k$}
        \State \Return $[c]\times k$
    \EndIf
    \State \Return \Call{RandomSample}{$\mathcal{C}$, k}
\EndFunction
\end{algorithmic}
\end{algorithm}
\end{minipage}

\end{table}    

\section{Experiments}
\label{sec:experiments}

\subsection{Dataset Creation}
\label{sec:dataset_creation}
To train the glyph generation model, we use an internal dataset of 15,000 fonts, including 40 million glyphs. We construct two versions of the dataset at 256×256 and 512×512 resolutions for two-stage training. For the texture expansion model, we curate a dataset of 542 PBR materials~\cite{FreePBR}, with 433 used for training and 109 for testing. We use Matplotlib~\cite{Hunter:2007} to render each glyph centered in a square image with a white background. For evaluating LoGAN, we focus on \textbf{word-level} assessment to assess complete font rendering rather than individual glyphs. 

Our evaluation dataset has two parts: (1) we collect 50 unseen fonts to create 300 synthetic movie-title pairs in the format (\emph{Movie1 in LanguageA}, \emph{Movie1 in LanguageB}), and we use 50 of the 300 pairs to add random textures sampled from the PBR materials test set; and (2) we collect 700 real font-based movie logos with corresponding language pairs in various monochrome colors. To construct the synthetic movie-title pairs, we collect movie titles and their translations—spanning Latin, Cyrillic, Greek, and CJK scripts—from the IMDb Top Movies list~\cite{IMDb}. We use this list to render the test fonts with random colors and textures (from the PBR test set) to increase the evaluation diversity. For the unseen test set, the selection criteria are as follows: (1) we choose fonts that were released substantially later than those in the training set (a gap of at least 1 year); (2) we compute similarity scores using SigLIP2 embeddings on the same set of 10 randomly sampled glyph images (A--Z), and keep only fonts whose average cosine similarity score is $< 0.5$.

\subsection{Quantitative Evaluation}
\begin{table}[t!]
\centering
\normalsize 
\caption{\textbf{Qualitative Comparison.} We measure PSNR, SSIM, LPIPS~\cite{zhang2018lpips}, DreamSim (DSim)~\cite{fu2023dreamsim}, and CLIP similarity
score~\cite{radford2021clip}. Results are shown for Non-CJK (254 pairs) and CJK (42 pairs). Editing type indicates whether the model is trained for text rerendering. Red/yellow denote 1st/2nd.}
\setlength{\tabcolsep}{8pt} 
\resizebox{\linewidth}{!}{
\begin{tabular}{l@{\hspace{3pt}}lcccccc} 
\toprule
\textbf{Lang. (\#)} & \textbf{Model} & \textbf{Editing Type} & \textbf{PSNR$\uparrow$} & \textbf{SSIM$\uparrow$} & \textbf{LPIPS$\downarrow$} &
\textbf{DSim$\downarrow$} & \textbf{CLIP$\uparrow$} \\
\midrule
\multirow{8}{*}{\rotatebox{90}{\hspace{0pt}CJK (42)\hspace{0pt}}}
& AnyText2~\cite{tuo2024anytext2} & Text & 6.894& 0.329& 0.825& 0.510& 0.811\\
& TextDiffuser-2~\cite{chen2024textdiffuser2} & Text & 6.460 & 0.449 & 0.882 & 0.604 & 0.666 \\
& Flux-Text~\cite{lan2025flux} & Text & 6.900& 0.580& 0.706& 0.374& 0.858\\
& Flux-Kontext Pro~\cite{bfl2025flux} & Image & 7.050& 0.635& 0.667& 0.162& 0.801\\
& Nano-Banana~\cite{google2025nanobanana} & Image & 7.658& 0.622& 0.588& 0.101& 0.920\\
& LoGAN w/o LoRA (GPT-5.2) & Text & \cellcolor{cellYellow}{9.052} & 0.657 & \cellcolor{cellYellow}{0.449} & 0.084 & \cellcolor{cellYellow}{0.932} \\
& LoGAN (Qwen) & Text & 8.186 & \cellcolor{cellYellow}{0.671} & 0.536 & \cellcolor{cellYellow}{0.081} & 0.929 \\
& LoGAN (GPT-5.2) & Text & \cellcolor{cellRed}{9.494}& \cellcolor{cellRed}{0.692}& \cellcolor{cellRed}{0.446}& \cellcolor{cellRed}{0.076}& \cellcolor{cellRed}{0.948}\\
\midrule
\multirow{8}{*}{\rotatebox{90}{\hspace{2pt}Non-CJK (254)\hspace{2pt}}}
& AnyText2~\cite{tuo2024anytext2} & Text & 7.123& 0.458& 0.806& 0.524& 0.812\\
& TextDiffuser-2~\cite{chen2024textdiffuser2} & Text & 6.875 & 0.541 & 0.862 & 0.631 & 0.658 \\
& Flux-Text~\cite{lan2025flux} & Text & 6.973& 0.560& 0.706& 0.444& 0.838\\
& Flux-Kontext Pro~\cite{bfl2025flux} & Image & 7.784& 0.596& 0.607& 0.158& 0.876\\
& Nano-Banana~\cite{google2025nanobanana} & Image & 8.988& 0.638& 0.481& \cellcolor{cellYellow}{0.084}& 0.922\\
& LoGAN w/o LoRA (GPT-5.2) & Text & 9.094 & 0.654 & \cellcolor{cellYellow}{0.444} & 0.085 & \cellcolor{cellYellow}{0.932} \\
& LoGAN (Qwen) & Text & \cellcolor{cellYellow}{9.162} & \cellcolor{cellYellow}{0.664} & 0.451 & 0.109 & 0.927 \\
& LoGAN (GPT-5.2) & Text & \cellcolor{cellRed}{10.077}& \cellcolor{cellRed}{0.684}& \cellcolor{cellRed}{0.373}& \cellcolor{cellRed}{0.072}& \cellcolor{cellRed}{0.944}\\
\bottomrule
\end{tabular}
}

\label{tab:results_category}
\end{table}

We compare two versions of LoGAN: one using \texttt{GPT-5.2} as the central VLM agent, and the other using the open-source Qwen3-VL-32B-Instruct model and compare against baselines specifically trained for texts (Flux-Text~\cite{lan2025flux}, AnyText2~\cite{tuo2024anytext2}), TextDiffuser-2~\cite{chen2024textdiffuser2}, as well as general generative models (Nano-Banana~\cite{google2025nanobanana}, FLUX-Kontext Pro~\cite{bfl2025flux}). We compute pixel-wise metrics PSNR and SSIM, alongside perceptual metrics LPIPS~\cite{zhang2018lpips}, DreamSim~\cite{fu2023dreamsim}, and CLIP Score~\cite{radford2021clip}.

Table~\ref{tab:results_category} presents our results. We observe substantial improvements across all pixel-wise and perceptual metrics. Interestingly, models specifically trained on text rendering perform worse than recent general-purpose image editing models. All baselines perform significantly worse when the output language is CJK, likely due to the scarcity of multilingual training data. In contrast, LoGAN (with both open-source and proprietary VLMs) benefits from the strong Glyph-Generation model, leading to comparable performance across metrics for both CJK and non-CJK outputs. We also report results for LoGAN without a fine-tuned LoRA using \texttt{GPT-5.2}, highlighting the benefit of style fine-tuning in reducing the domain gap and improving overall style transfer quality while \textbf{requiring only the reference image itself}. Notably, LoGAN still outperforms the baselines even without LoRA.

\subsection{Qualitative Evaluation}
 
\begin{figure*}[h!] 
    \centering 
    \includegraphics[width=0.99\linewidth]{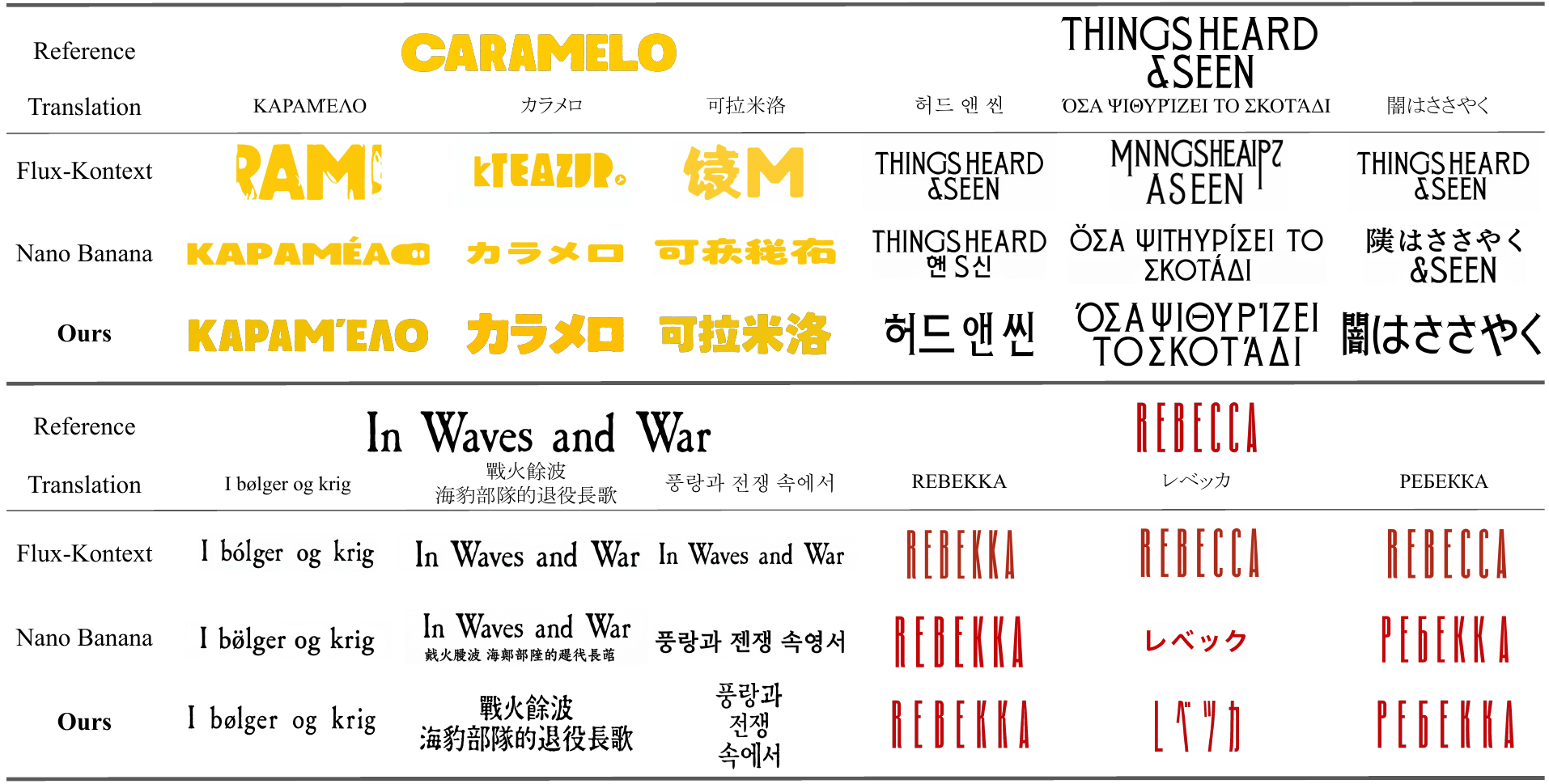}
    \caption{\textbf{Qualitative Comparison.} We present the results of 12 localized logos spanning across 4 movie titles. Flux-Kontext often ignores the text guidance and does not perform any edits, especially for CJK languages. Nano-Banana often misses the translations and slightly alters the style. Our results performs accurate style and layout transfer while simultaneously translating the titles correctly. } 
    \label{fig:qualitative_comparison} 
    \vspace{-0.1in}
\end{figure*}

\noindent\textbf{Font Quality Comparison.}
Figure~\ref{fig:qualitative_comparison} presents 12 examples of 12 movies logos from four titles. We compare LoGAN (\texttt{GPT-5.2}) against two strong baselines: Nano-Banana~\cite{google2025nanobanana} and FLUX Kontext Pro~\cite{bfl2025flux}. FLUX Kontext Pro often produces completely illegible results (e.g., all instances of "Caramelo") or fails to perform any edits (columns 1 and 3 for "Things Heard \& Seen" and columns 2 and 3 for “In Waves and War”). Nano-Banana, while performing better in English (column 1 for "Rebecca"), frequently misses diacritics in other Latin-based languages (column 1 for "In Waves and War"), slightly changes the style (all instances of "Caramelo"), or produces incorrect translations for non-Latin scripts (column 2 for “Things Heard \& Seen” and "In Waves and War"). In contrast, LoGAN accurately preserves the layout, spacing, and style of the original title and applies them consistently to both CJK and non-CJK languages. Column 2 of "Rebecca" is a clear example, where the tall, slim style is captured and correctly transferred to Japanese. \\

\begin{figure}[ht!]
    \centering
    \includegraphics[width=\linewidth]{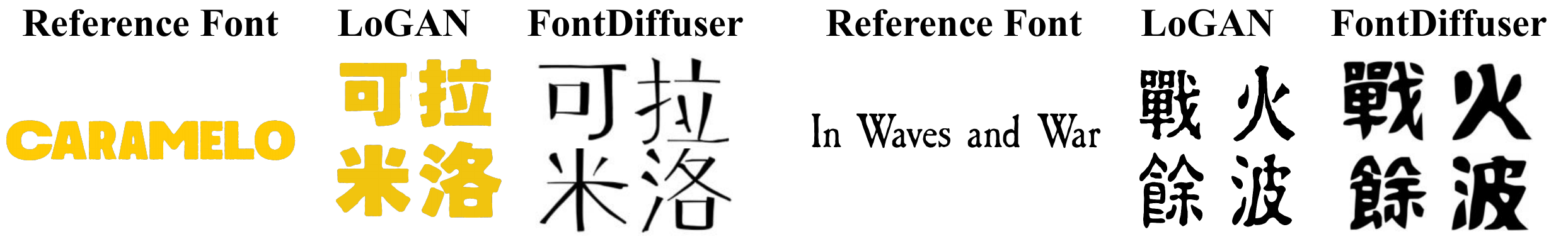}
    \caption{\textbf{Glyph Quality Comparison Against FontDiffuser}. FontDiffuser fails to match the original style/color of the reference font. 
    }
    \label{fig:additional_results}
\end{figure}

\noindent\textbf{Glyph Quality Comparison.}
Existing font-based methods, such as FontDiffuser~\cite{yang2024fontdiffuser}, cannot predict spacing/kerning and align multiple characters. We therefore provide an additional qualitative comparison to highlight LoGAN's glyph generation quality. Since FontDiffuser only accepts and generates a single character, we use randomly cropped glyphs from a logo as references when generating characters. We show comparisons in Chinese (their strongest language) in Figure~\ref{fig:additional_results}. LoGAN produces more accurate styles and colors. \\

\noindent\textbf{Texture Expansion Quality Comparison.}
We also perform a qualitative comparison of our Texture-Expansion model. To the best of our knowledge, we are the first to train a model that recovers texture from a given text image with a burnt-in texture. Thus, our closest comparison is a general image editing model prompted with \texttt{Recover the texture from the given text image}.

  \begin{figure}[h] 
      \centering
      \includegraphics[width=0.99\linewidth]{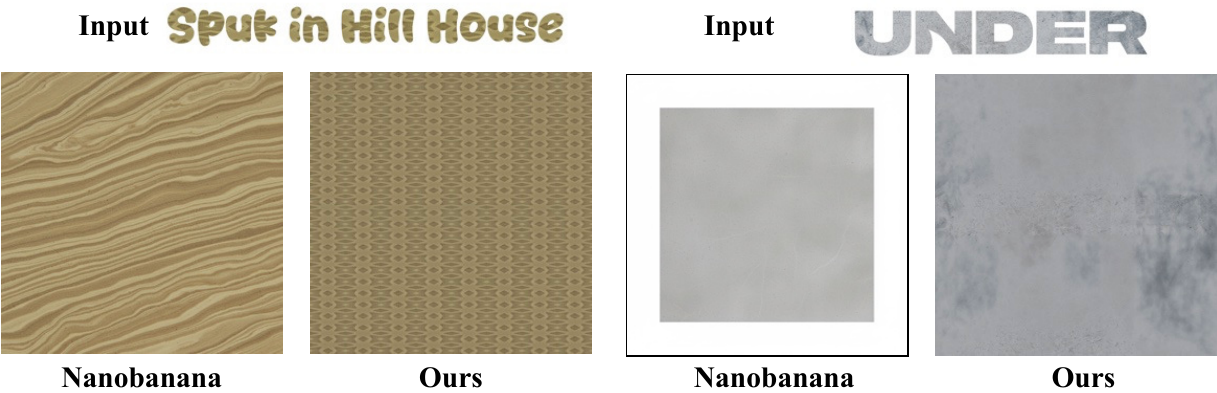}
      \caption{\textbf{Texture Expansion Qualitative Comparison.} We present the qualitative results of our texture expansion approach against
  Nano-Banana~\cite{google2025nanobanana}. }
      \label{fig:texture}

  \end{figure}
Figure~\ref{fig:texture} shows qualitative comparisons between our model and Nano-Banana~\cite{google2025nanobanana}. Since Nano-Banana is not specifically trained for texture expansion, it often generates incorrect textures (left) or loses fine-grained details (right). In contrast, our model better preserves fine details and more closely matches the original texture in the reference inputs.

\subsection{Expert Evaluation}
To better assess fine-grained details in generated fonts of LoGAN, we worked with a font designer rather than conducting a user study, because non-expert users often \textbf{overlook the intricate details of font design}. Specifically, the designer was asked to evaluate the generated fonts in terms of: (1) style transfer (whether the generated letters match the reference’s visual styling, e.g., stroke shape/weight, curvature, color, and texture), (2) kerning/spacing (between adjacent characters), and (3) translation accuracy (whether the generated glyphs match the prompt). The evaluation had two parts: (1) evaluating 80 generated fonts (from five randomly selected reference movie titles), comparing LoGAN against the strongest baseline, Nano-Banana, using a 3-point scale (1, 2, 3), where 1 = completely inaccurate, 2 = decent but with minor issues, and 3 = completely accurate and ready to use; and (2) asking the designer to classify all 700 generated font-based logos as strictly \textbf{"Production-Ready"} or not.

In part 1, LoGAN significantly outperformed the baseline, achieving average scores of 2.8 for style transfer and 2.2 for layout/spacing. Nano-Banana performed poorly (average 1.0) in both categories, underscoring the difficulty of high-fidelity layout preservation. In addition, LoGAN was consistently accurate in translation (average 3.0), whereas Nano-Banana translated the logos correctly only 19 out of 80 times. For part 2, 52.9\% of LoGAN’s outputs were classified as Production-Ready, demonstrating the strong generalization of our method. Notably, none of Nano-Banana’s logos passed this test.

\subsection{More Ablation Studies}

\subsubsection{Single-Glyph Analysis.}
To mitigate the potential effects of positional discrepancies (e.g., kerning), we compare LoGAN against the strongest baseline, Nano-Banana, by cropping all generated logos into individual glyphs using tight crops to avoid spacing bias and evaluating them with the same five metrics as in the main paper. The results are presented in Table~\ref{tab:glyph_metrics}, showing that LoGAN outperforms Nano-Banana at the glyph level.

\begin{table}[h]   
\centering                             
\small                    
\caption{Glyph-level comparison for LoGAN and Nano-Banana.}
\setlength{\tabcolsep}{5pt}
\begin{tabular}{lccccc}            
\toprule 
Method & PSNR$\uparrow$ & SSIM$\uparrow$ & LPIPS$\downarrow$ & DreamSim$\downarrow$ & CLIP$\uparrow$ \\               
\midrule            
NB & 15.75 & 0.726 & 0.179 & 0.122 & 0.958 \\
LoGAN        & \textbf{18.67} & \textbf{0.802} & \textbf{0.121} & \textbf{0.094} & \textbf{0.968} \\  
\bottomrule 
\end{tabular} 
\label{tab:glyph_metrics}
\end{table}

\subsubsection{Texture Expansion Analysis.}
To further validate the impact of the developed Texture-Expansion model, we conducted additional evaluations. We split the 300 test pairs into two groups (see more details in Section~\ref{sec:dataset_creation}): \emph{with texture} (50) and \emph{without texture} (250). The results in Table~\ref{tab:image_metrics_split} show that LoGAN with the Texture-Expansion model achieves better performance than state-of-the-art methods, such as Nano-Banana.

\begin{table}[h]
\centering                                   \small    
\setlength{\tabcolsep}{3pt}
\caption{Quantitative comparisons for LoGAN and Nano-Banana in texture/no texture split.} 
\begin{tabular}{llccccc}
\toprule
Group & Method & PSNR$\uparrow$ & SSIM$\uparrow$ & LPIPS$\downarrow$ & DSim$\downarrow$ & CLIP$\uparrow$ \\
\midrule  
\multirow{2}{*}{With Texture}
& Nano-Banana   & 10.65 & 0.606 & 0.483 & \textbf{0.100} & 0.916 \\        
& LoGAN & \textbf{11.88} & \textbf{0.659} & \textbf{0.378} & 0.118 & \textbf{0.934} \\
\midrule
\multirow{2}{*}{Withou texture}
& Nano-Banana   & 7.85 & 0.647 & 0.504 & 0.081 & 0.924 \\ 
& LoGAN & \textbf{9.00} & \textbf{0.699} & \textbf{0.391} & \textbf{0.049} & \textbf{0.951} \\ 
\bottomrule       
\end{tabular}                             
\label{tab:image_metrics_split}
\end{table}

\subsubsection{Efficiency of LoGAN.}
We benchmark LoGAN’s runtime (3 runs, \texttt{GPT-5.2}) in two settings:
(i) Stage 1: design extraction on a fixed-size synthetic reference image while varying the rendered text length from 1 to 15 complex CJK characters (the average length of a movie title); and
(ii) Stages 2--3: generation with 30 denoising steps while varying the output length from 1 to 15 characters.
In our experiments, Stage 1 runtimes range from 7.6\,s (1 char) to 38.5\,s (15 chars).
Stages 2--3 take 25.5\,s (1 char) to 43.2\,s (15 chars) on an A100 (40GB).
LoRA finetuning incurs a one-time cost of about 15 minutes.
We emphasize that our goal is not to optimize baseline speed but to address their quality limitations.

\subsubsection{VLM for Detecting Font Styles.} \label{sec:supp_vlm}
In Stage 1 of our pipeline (Figure~\ref{fig:LoGAN_design_overview}), we use \texttt{GPT-5.2} (VLM) to identify font styles and represent them in an abstract form. To evaluate the robustness of this approach, we collected a new dataset of diverse movie logos from 400 titles released in 2025 across multiple languages. We asked a font designer to group them into three categories (1, 2, 3 font styles or more complex designs beyond font) and asked \texttt{GPT-5.2} to classify the number of unique font styles in each sample. Our results show that \texttt{GPT-5.2} achieves 93.1\% accuracy, validating that our VLM-based approach can robustly interpret diverse font styles and guide the subsequent tasks of LoGAN.

\subsubsection{Pipeline Robustness Test.} We also evaluated LoGAN’s end-to-end pipeline robustness, which is crucial for production use. Here, we do not focus on assessing model quality. Instead, we evaluate whether the pipeline can consistently generate meaningful outputs (e.g., complete rendered texts). To this end, we conduct a large-scale pressure test on a different dataset of 1,800 movie (font-based) logo pairs (e.g., English vs other languages). We run LoGAN (with \texttt{GPT-5.2}) end-to-end on this dataset. The results show that LoGAN is highly robust, with only a 3\% pipeline failure rate across 1,800 runs. Failures are primarily due to (i) API responses not conforming to the required JSON schema, or (ii) VLM errors (e.g., incorrect character detection or design transfer on minority languages), which lead to downstream failures.

\section{Conclusion}
\label{sec:conclusion}
We present LoGAN, a VLM-based agentic framework for end-to-end font localization. Prior work often lacks multilingual support and complete font rendering. LoGAN addresses these limitations by proposing several novel components, including a cross-language sampling algorithm, a multilingual glyph generation model, character-level text detection, a spacing and kerning transfer algorithm, and a texture expansion model. Our empirical results show that LoGAN outperforms font generation and state-of-the-art image editing baselines, and our pressure test on the end-to-end pipeline further indicates that LoGAN is a robust solution for real-world practical use.

In future work, we plan to improve LoGAN along three directions: (1) extending beyond single-font localization to support multi-font rendering and more flexible layout and arrangement; (2) improving fine-tuning efficiency—while fine-tuning is optional in our current design, making it faster and more lightweight would improve practicality; and (3) supporting more complex font/logo rendering. Currently, LoGAN relies on glyph-based generation and may struggle with connected characters, e.g., calligraphy fonts.

\clearpage  


%
%
\bibliographystyle{splncs04}
\bibliography{main}

\newpage
\begin{center}
\LARGE \textbf{LoGAN: Supplementary Material}
\end{center}

\setcounter{page}{1}
In this supplement, we provide additional details about LoGAN. We present detailed model configurations in Sections~\ref{sec:supp_model_config}  on the Glyph-Generation and Texture-Expansion models and their respective LoRA fine-tuning layers. We also describe the prompts used for various VLM-related tasks in Section~\ref{sec:supp_prompts}, as well as the list of supported languages for LoGAN in Section~\ref{sec:supp_languages}.

We also provide more empirical studies on PNG-to-SVG conversion quality (Section~\ref{sec:supp_svg}), additional texture expansion results (Section~\ref{sec:supp_texture_expansion}), qualitative results (Section~\ref{sec:supp_additional_quals}), and quantitative comparisons against baselines, including Nano-Banana Pro (Section~\ref{sec:supp_quant_comparison}).

\section{Model Configuration}
\label{sec:supp_model_config}
We summarize the model configuration for the Glyph-Generation model in Table~\ref{tab:glyphgen_model_config}. The model is similar to \texttt{SD3.5-medium}\cite{stabilityai_stable_diffusion_35_medium}, with modifications to the joint attention layer to support an additional image input modality, as introduced in Section\ref{sec:texture}. The model architecture implementation is based on \texttt{transformer\_sd3.py} in the \texttt{diffusers} library~\cite{von-platen-etal-2022-diffusers}. We do not use pretrained weights. Instead, we train the model from scratch.

For Texture-Expansion model, we use PEFT~\cite{peft} to implement LoRA fine-tuning, as shown in Figure~\ref{lst:lora_texture_expansion}, based on the \texttt{FLUX.1-Fill-dev} model~\cite{blackforestlabs_flux1_fill_dev}. For style fine-tuning, we use PEFT to implement LoRA layers, as shown in Figure~\ref{lst:lora_glyph_generation}, based on the pretrained Glyph-Generation model.

\begin{table}[h] 
\centering
\small
\setlength{\tabcolsep}{5pt}
\caption{Configurations for Glyph-Generation model.}
\begin{tabular}{@{}ll@{}}
\toprule
\textbf{Parameter} & \textbf{Value} \\
\midrule
\multicolumn{2}{@{}l}{\textit{Noise Scheduler}} \\
Shift & 1.0 (3.0 for 512x512) \\
Timesteps & 1000 \\
\midrule
\multicolumn{2}{@{}l}{\textit{Transformer (SD3Transformer2DModel)}} \\
Sample size & 128 \\
Patch size & 2 \\
Channels (in/out) & 16 / 16 \\
Layers & 18 \\
Attention heads & 24 \\
Head dimension & 48 \\
Attention dim (joint) & 1152 \\
Caption dimension & 1536 \\
Positional embed max size & 384 \\
QK normalization & rms\_norm \\
Dual attention layers & [0--9] \\
Resampler depth & 6 \\
Resampler heads & 20 \\
Resampler head dim & 64 \\
Resampler queries & 256 \\
\midrule
\multicolumn{2}{@{}l}{\textit{VAE}} \\
Configuration & AuraDiffusion/16ch-vae \\
\midrule
\multicolumn{2}{@{}l}{\textit{Image Encoder}} \\
Model & siglip2-so400m-patch16-256 \\
Image size & 256 \\
Embedding dimension & 1152 \\
\midrule
\multicolumn{2}{@{}l}{\textit{Text Encoder}} \\
Model & google/byt5-large \\
Embedding dimension & 1536 \\
\bottomrule
\end{tabular}
\label{tab:glyphgen_model_config}
\end{table}

\section{SVG Conversion Quality}
\label{sec:supp_svg}
Motivated by~\cite{li2024hfh}, we observe that, with high-quality glyph images, the quality loss from raster images to vectors is negligible. Therefore, we adopt a similar strategy: instead of training an end-to-end SVG generation model, we select a robust PNG-to-SVG tool to vectorize our image outputs and make them editable. For our experiments, we generated 100 random glyphs (512×512 images) sampled from our font test set across multiple languages. We convert these glyphs to SVG using several PNG-to-SVG approaches—including commercial tools such as ImageTrace (Adobe Illustrator) and VectorMagic~\cite{cedarlake_vectormagic}, as well as open-source tools such as Potrace~\cite{selinger_potrace_2019} and StarVector-8B~\cite{rodriguez2024starvector}—and asked a professional font designer to evaluate the outputs and rank the results.
 
\begin{figure}[h] 
  \centering
  \includegraphics[width=0.7\columnwidth]{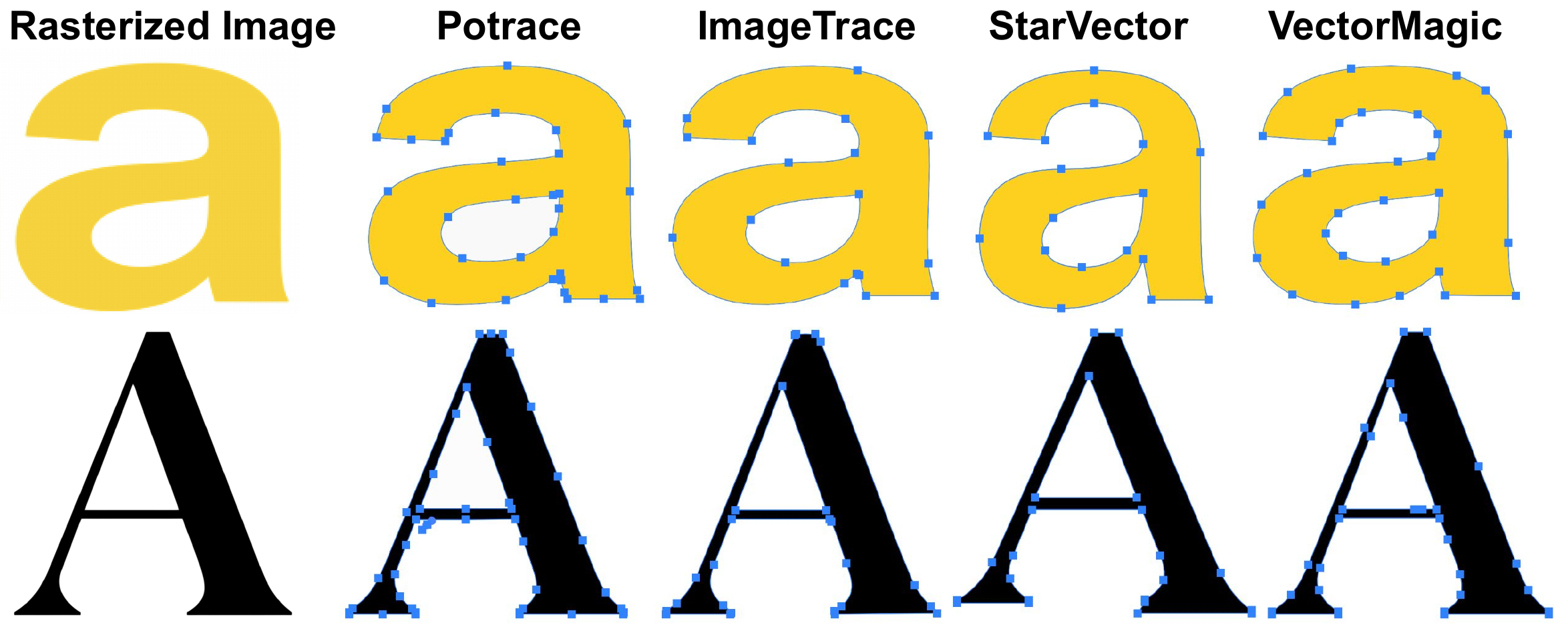}
\caption{\textbf{Comparison of PNG to SVG quality using different approaches}. The leftmost column shows the rasterized image (PNG), while the right columns display results from Potrace, ImageTrace, StarVector (8B), and VectorMagic. Note that colors may have shifted slightly due to the screenshot and copy-paste from the source file.}
\label{fig:svg_comparison}
\end{figure}

Overall, the evaluation shows that VectorMagic consistently performs better than the other three approaches. Thus, we primarily use it in our pipeline. Among the open-source solutions, Potrace performs consistently better than StarVector-8B. We present two representative examples for illustration in Figure~\ref{fig:svg_comparison}. This comparison shows that: (1) non-generative approaches are more stable and produce fewer variations; and (2) VectorMagic produces the cleanest and most accurate vector outputs.

\begin{figure}[h]
\begin{lstlisting}[style=mypythonstyle, basicstyle=\scriptsize\ttfamily, xleftmargin=1pt, xrightmargin=1pt]
lora_targets = [
    # Attention layers
    "attn.add_k_proj", "attn.add_q_proj", 
    "attn.add_v_proj", "attn.to_add_out",
    "attn.to_k", "attn.to_q", "attn.to_v",
    "attn.to_out.0",
    "attn2.to_k", "attn2.to_q", "attn2.to_v",
    "attn2.to_out.0",
    "attn.processor.to_{q,k,v}_ip",
    "attn2.processor.to_{q,k,v}_ip",
    
    # Feed-forward & norms
    "ff.net.0.proj", "ff.net.2",
    "ff_context.net.0.proj", "ff_context.net.2",
    "norm1.linear", "norm1_context.linear",
    "attn*.processor.norm_ip.linear",

    # Image projection
    "image_proj_model.proj_{in,out}",
    "image_proj_model.layers.*.attn.to_q",
    "image_proj_model.layers.*.attn.to_kv",
    "image_proj_model.layers.*.attn.to_out.0",
    "image_proj_model.layers.*.ff.net.*",
    "image_proj_model.layers.*.adaln_proj",
]
\end{lstlisting}
\caption{LoRA Target Layers for Style LoRA Fine-tuning on Glyph-Generation Model.}
\label{lst:lora_glyph_generation}
\end{figure}

\begin{figure}[h]
\begin{lstlisting}[style=mypythonstyle, basicstyle=\scriptsize\ttfamily, xleftmargin=1pt, xrightmargin=1pt]
lora_targets = [
    # Attention layers
    "attn.to_k", "attn.to_q", "attn.to_v",
    "attn.add_k_proj","ttn.add_q_proj","attn.add_v_proj",
    "attn.to_add_out",
    
    # Feed-forward & norms
    "ff.net.0.proj", "ff.net.2",
    "ff_context.net.0.proj", "ff_context.net.2"
]
\end{lstlisting}
\caption{LoRA target layers for Texture-Expansion Model.}
\label{lst:lora_texture_expansion}
\end{figure}

\section{List of Supported Languages}
\label{sec:supp_languages}
To the best of our knowledge, this is the \textbf{first glyph generation model to support such broad language coverage}. The languages supported by LoGAN include: Bokmål, Chinese Simplified, Chinese Traditional, Czech, Danish, Dutch, Finnish, French, French-Canadian, German, Greek, Hungarian, Indonesian, Italian, Japanese, Korean, Polish, Portuguese, Portuguese (Brazilian), Romanian, Russian, Spanish (Castilian), Spanish (Neutral), Swedish, Turkish, Vietnamese, Ukrainian, and Filipino. 

\section{List of Prompts}
\label{sec:supp_prompts}
We present the detailed prompts used in different stages of the LoGAN pipeline. The stage 1 prompt for extracting design attributes is presented in Figure~\ref{fig:prompt_design_extraction}, the stage 2 prompt used for transferring design attributes to target text is presented in Figure~\ref{fig:prompt_design_transfer}. The prompt used for glyph cropping (AutoCrop) to individual letters is presented in Figure~\ref{fig:prompt_auto_cropping}.

\section{Baseline Implementations}
\label{sec:supp_baselines}
We briefly describe the baseline implementations. For AnyText2, we use the entire image as mask of canvas, and feed the model with our reference logo on top of a white background with the text prompt of \texttt{Change the text to [Target Text]}. Flux-Text and TextDiffuser-2 both require an additional image hint on where and what texts should be placed. Thus, we take an extra step of masking out the the texts in the reference title, then render a default text of the translation in the masked part. We then feed this image hint, along with mask, reference logo, and the same text prompt into the model. Flux-Kontext Pro and Nano-Banana comparisons are both done via API calls with the prompt \texttt{Change the text to [Target Text] while keeping the original style}.

\section{Additional Quantitative Comparisons}
\label{sec:supp_quant_comparison}
Given the new release of Nano-Banana Pro, we provide additional comparisons against it. Table~\ref{tab:results_category2} presents our results—\methodname{} still outperforms it on both CJK and non-CJK languages, despite Nano-Banana Pro being significantly better than its previous version at rendering these languages.

\section{Additional Qualitative Results}
\label{sec:supp_additional_quals}
\begin{figure*}[ht!] 
    \centering 
    \includegraphics[width=\textwidth]{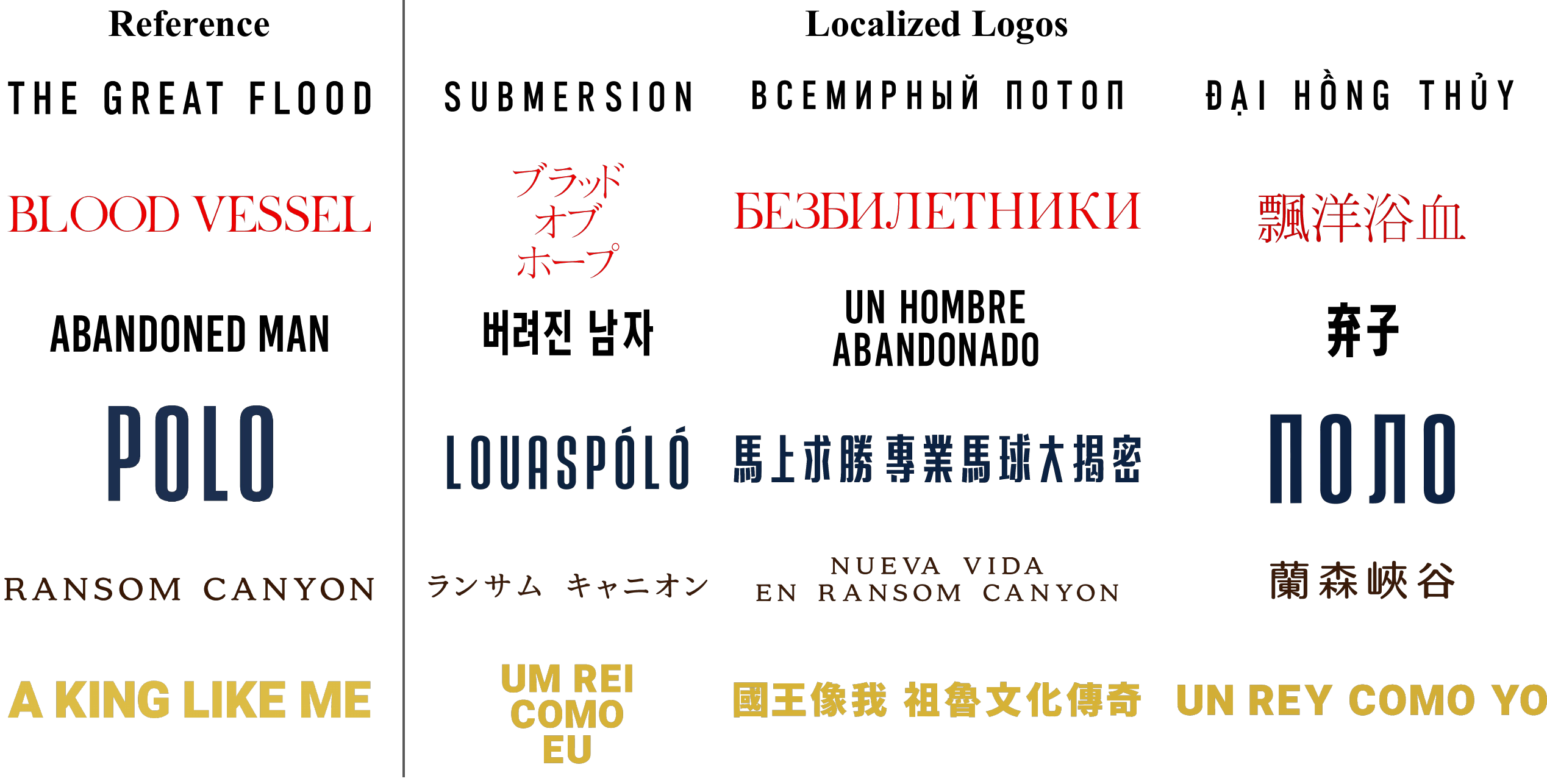} 
    \caption{\textbf{Qualitative Results.} We present 15 more generated results from \methodname{} across 5 titles from from recently released movies.} 
    \label{fig:supp_quals} 
\end{figure*}

Figure~\ref{fig:supp_quals} presents 15 additional results from \methodname{}. All examples show that LoGAN precisely and accurately transfers the font style, color, and texture from the source language (reference image) to target languages (generated image).

\begin{table*}[h!]
\centering
\normalsize 
\caption{\textbf{Qualitative Comparison.} We measure PSNR, SSIM, LPIPS~\cite{zhang2018lpips}, DreamSim (DSim)~\cite{fu2023dreamsim}, and CLIP similarity
score~\cite{radford2021clip}. Results are shown for Non-CJK (254 pairs) and CJK (42 pairs). Editing type indicates whether the model is trained for text rerendering. Red/yellow denote 1st/2nd.}
\setlength{\tabcolsep}{8pt} 
\resizebox{\linewidth}{!}{
\begin{tabular}{l@{\hspace{3pt}}lcccccc} 
\toprule
\textbf{Lang. (\#)} & \textbf{Model} & \textbf{Editing Type} & \textbf{PSNR$\uparrow$} & \textbf{SSIM$\uparrow$} & \textbf{LPIPS$\downarrow$} &
\textbf{DSim$\downarrow$} & \textbf{CLIP$\uparrow$} \\
\midrule
\multirow{8}{*}{\rotatebox{90}{\hspace{0pt}CJK (42)\hspace{0pt}}}
& AnyText2~\cite{tuo2024anytext2} & Text & 6.894& 0.329& 0.825& 0.510& 0.811\\
& TextDiffuser-2~\cite{chen2024textdiffuser2} & Text & 6.460 & 0.449 & 0.882 & 0.604 & 0.666 \\
& Flux-Text~\cite{lan2025flux} & Text & 6.900& 0.580& 0.706& 0.374& 0.858\\
& Flux-Kontext Pro~\cite{bfl2025flux} & Image & 7.050& 0.635& 0.667& 0.162& 0.801\\
& Nano-Banana~\cite{google2025nanobanana} & Image & 7.658& 0.622& 0.588& 0.101& 0.920\\
& Nano-Banana Pro~\cite{google2025nanobanana} & Image & \cellcolor{cellYellow}{9.055} & 0.635 & \cellcolor{cellRed}{0.429} & 0.082 & \cellcolor{cellRed}{0.949}\\
& LoGAN w/o LoRA (GPT-5.2) & Text & 9.052 & 0.657 & 0.449 & 0.084 & 0.932 \\
& LoGAN (Qwen) & Text & 8.186 & \cellcolor{cellYellow}{0.671} & 0.536 & \cellcolor{cellYellow}{0.081} & 0.929 \\
& LoGAN (GPT-5.2) & Text & \cellcolor{cellRed}{9.494}& \cellcolor{cellRed}{0.692}& \cellcolor{cellYellow}{0.446}& \cellcolor{cellRed}{0.076}& \cellcolor{cellYellow}{0.948}\\
\midrule
\multirow{8}{*}{\rotatebox{90}{\hspace{2pt}Non-CJK (254)\hspace{2pt}}}
& AnyText2~\cite{tuo2024anytext2} & Text & 7.123& 0.458& 0.806& 0.524& 0.812\\
& TextDiffuser-2~\cite{chen2024textdiffuser2} & Text & 6.875 & 0.541 & 0.862 & 0.631 & 0.658 \\
& Flux-Text~\cite{lan2025flux} & Text & 6.973& 0.560& 0.706& 0.444& 0.838\\
& Flux-Kontext Pro~\cite{bfl2025flux} & Image & 7.784& 0.596& 0.607& 0.158& 0.876\\
& Nano-Banana~\cite{google2025nanobanana} & Image & 8.988& 0.638& 0.481& 0.084 & 0.922\\
& Nano-Banana Pro~\cite{google2025nanobanana} & Image & \cellcolor{cellYellow}{9.336} & 0.611 & \cellcolor{cellYellow}{0.392} & \cellcolor{cellYellow}{0.084} & \cellcolor{cellYellow}{0.944} \\
& LoGAN w/o LoRA (GPT-5.2) & Text & 9.094 & 0.654 & 0.444 & 0.085 & 0.932 \\
& LoGAN (Qwen) & Text & 9.162 & \cellcolor{cellYellow}{0.664} & 0.451 & 0.109 & 0.927 \\
& LoGAN (GPT-5.2) & Text & \cellcolor{cellRed}{10.077}& \cellcolor{cellRed}{0.684}& \cellcolor{cellRed}{0.373}& \cellcolor{cellRed}{0.072}& \cellcolor{cellRed}{0.944}\\
\bottomrule
\end{tabular}
}

\label{tab:results_category2}
\end{table*}

\section{Additional Texture Expansion}
\label{sec:supp_texture_expansion}
Figure~\ref{fig:supp_quals} presents six additional results from the test dataset, as mentioned in Section~\ref{sec:experiments}, using our texture expansion model. Our model is able to recover both simple and complex textures from the reference images. Surprisingly, even when the strokes are very thin, the model can still pick up these cues and produce convincing texture expansion results.

\begin{figure*}[ht!] 
    \centering 
    \includegraphics[width=\textwidth]{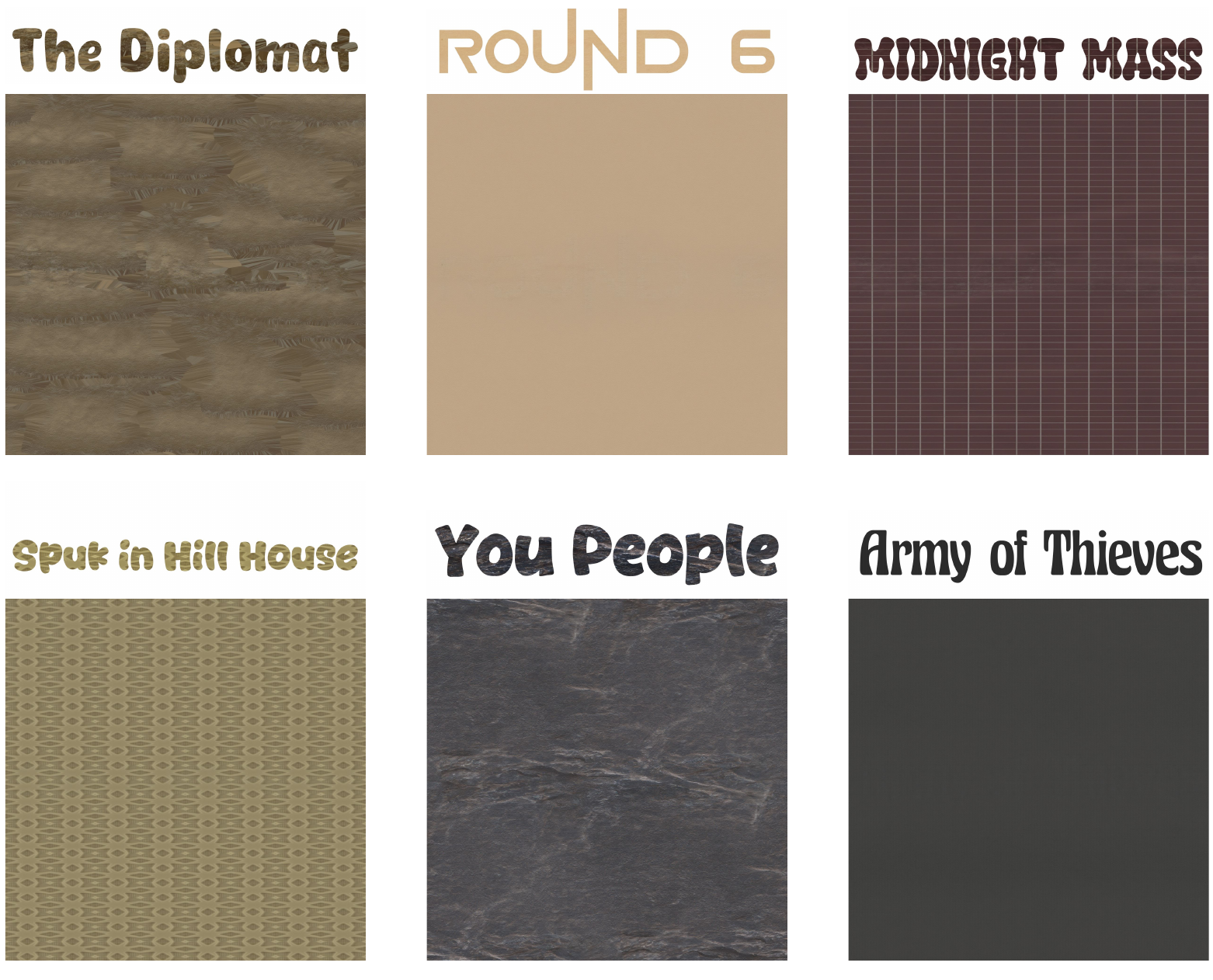} 
    \caption{\textbf{Texture Expansion Results.} We present more results of texture expansion from our synthetic testing dataset. Our model is able to extract detailed structures even with very little hint from the given image (e.g., "Lion King" example where the strokes are extremely thin). Zoom in for details on the textures.} 
    \label{fig:supp_texture} 
\end{figure*}

\begin{figure*}[t] 
\centering
\begin{mybox}{Prompt for Design Extraction}
\textbf{Role:} You are an font/logo analyzer.

\textbf{Task:} From the provided font/logo image, perform character-level OCR, group characters into abstract visual styles, extract per-character colors, and analyze overall letter case.

\textbf{Process:}
\begin{itemize}
    \setlength\itemsep{0em} 
    \item Run OCR to extract all text characters from the logo.
    \item Build a sequential list of characters (exclude empty spaces) and number them starting at 1.
    \item For each character:
    \begin{itemize}
        \item Assign a color in hex (e.g., \#RRGGBB). If color cannot be determined accurately, use ``unknown''.
        \item Assign an abstract style label based on visual similarity.
        \item Assign an abstract texture label based on texture of logo.
    \end{itemize}
    \item Detect letter case for the logo from one of these options: \texttt{all\_uppercase}, \texttt{mix\_upper\_lower\_case}, \texttt{all\_lowercase}
    \begin{itemize}
        \item If no text is detected, use \texttt{no\_text\_detected}.
        \item If text contains only numbers/symbols with no letters, use \texttt{no\_letters\_detected}.
    \end{itemize}
    \item Count the number of lines for the logo.
    \item Detect line breaks from the logo and return the index where each line break is applied. For example, for ``Hello\textbackslash nWorld'', return [5].
\end{itemize}

\textbf{Style grouping criteria:}
\begin{itemize}
    \setlength\itemsep{0em}
    \item Font family/typeface (same font but different color = same style)
    \item Font size (same style but different size = different styles)
    \item Font weight/thickness; Font angle
    \item Symbols/punctuation are included as characters but must use style ``unknown''
\end{itemize}

\textbf{Requirements:}
\begin{itemize}
    \setlength\itemsep{0em}
    \item Follow reading order strictly (top-to-bottom, then left-to-right).
    \item Exclude empty spaces from the character list.
    \item Include symbols/punctuation as characters; style must be ``unknown''.
    \item Use generic style/texture names (e.g., ``style1'', ``texture1''). No descriptive terms.
    \item Use standard hex color codes; if undetermined, use ``unknown''.
    \item If the logo contains only texture, use ``unknown'' for color, and vice versa.
    \item Do not include line break symbols in \texttt{extracted\_text}.
    \item Add kerning to the output JSON if input provided; otherwise empty.
    \item Output must be valid JSON only. No explanations.
    \item For duplicate characters, append suffix starting with ``\_1''.
    \item Handle multi-line spacing by removing space between last char of current line and first of next.
\end{itemize}

\textbf{Example Output (JSON):}
\begin{verbatim}
{
  "extracted_text": "[text detected without line break symbols]",
  "letter_case": "[all_uppercase|mix_upper_lower_case|...]",
  "num_lines": 0,
  "alignment": "[left|center|right]",
  "confidence": "[high|medium|low]",
  "kerning": {"H-E": 10, "E-L": 10, ...},
  "line_breaks": [5],
  "num_styles": 1,
  "style_groups": {
    "H": { "style": "style1", "color": "#000000", "texture": "unknown" },
    "E": { "style": "style1", "color": "#000000", "texture": "unknown" },
    "L": { "style": "style1", "color": "#000000", "texture": "unknown" },
    "L_1": { "style": "style1", "color": "#000000", "texture": "unknown" },
    "O": { "style": "style1", "color": "#000000", "texture": "unknown" }
  }
}
\end{verbatim}
\end{mybox}
\caption{Prompt for Design Extraction.}
\label{fig:prompt_design_extraction}
\end{figure*}

\begin{figure*}[t]
\centering
\begin{mybox}{Prompt for Design Transfer}
\textbf{Role:} You are a professional font/logo designer.

\textbf{Task:} Your task is to analyze a reference text with associated color/texture and style information and apply those colors/textures and styles to a target text on a character-by-character basis.

\textbf{Rules:}
\begin{itemize}
    \setlength\itemsep{0em}
    \item \textbf{Character-by-Character Mapping:} For each character in the \texttt{target} text, assign it a color/texture and style based on the \texttt{reference} style information. Ignore empty spaces in the target and reference text. Preserve case sensitivity on target text when mapping characters.
    \item \textbf{Duplicate Character Suffixes:} If a character appears more than once in the target text, append a numerical suffix to its key. The first instance has no suffix, the second gets \texttt{\_1}, the third gets \texttt{\_2}, and so on (e.g., `c', `c\_1', `c\_2').
    \item \textbf{Semantic Color/texture and Style Grouping:} This is the most important rule. When the reference text contains multiple words or phrases with different colors/texture and styles, you must identify the corresponding semantic equivalent in the target text (e.g., a translation of reference text). All characters within the target's translated word/phrase must inherit both the color/texture AND style of the original reference word/phrase.
    \item \textbf{Style Constraint:} You can ONLY use styles that exist in the reference style dictionary. Do not create or assign any styles beyond those provided in the reference.
\end{itemize}

\textbf{Input:}
\begin{itemize}
    \setlength\itemsep{0em}
    \item \texttt{reference}: A dictionary containing \texttt{style\_group} information and the original \texttt{title\_text}.
    \item \texttt{target}: The string to be colorized and styled.
\end{itemize}

\textbf{Output:}
\begin{itemize}
    \setlength\itemsep{0em}
    \item Return a single dictionary and nothing else.
    \item Do not provide any explanations, introductory text, or markdown formatting.
    \item Keys must be the characters from the target text (with suffixes for duplicates, if any).
    \item Values must be dictionaries containing both \texttt{color} (hex code as string) or \texttt{texture} and \texttt{style} (abstract style name from reference only).
    \item Outputs must contain all characters in Target text.
\end{itemize}

\textbf{Example:}
{\tiny
\begin{verbatim}
Reference:
{
  "style_group": {
    "D": {"color": "#5D2A1C", "texture": "unknown", "style": "style1"},
    "R": {"color": "#5D2A1C", "texture": "unknown", "style": "style1"},
    "I": {"color": "#5D2A1C", "texture": "unknown", "style": "style1"},
    "N": {"color": "#5D2A1C", "texture": "unknown", "style": "style1"},
    "K": {"color": "#5D2A1C", "texture": "unknown", "style": "style1"},
    "M": {"color": "#9b9a99", "texture": "unknown", "style": "style1"},
    "A": {"color": "#9b9a99", "texture": "unknown", "style": "style1"},
    "S": {"color": "#9b9a99", "texture": "unknown", "style": "style1"},
    "T": {"color": "#9b9a99", "texture": "unknown", "style": "style1"},
    "E": {"color": "#9b9a99", "texture": "unknown", "style": "style1"},
    "S_1": {"color": "#9b9a99", "texture": "unknown", "style": "style1"}
  },
  "title_text": "DRINK MASTERS"
}

Target: "Bậc thầy pha chế"
Output:
{
  "B": {"color": "#9b9a99", "texture": "unknown", "style": "style1"},
  "ậ": {"color": "#9b9a99", "texture": "unknown", "style": "style1"},
  "c": {"color": "#9b9a99", "texture": "unknown", "style": "style1"},
  "t": {"color": "#9b9a99", "texture": "unknown", "style": "style1"},
  "h": {"color": "#9b9a99", "texture": "unknown", "style": "style1"},
  "ầ": {"color": "#9b9a99", "texture": "unknown", "style": "style1"},
  "y": {"color": "#9b9a99", "texture": "unknown", "style": "style1"},
  "p": {"color": "#5D2A1C", "texture": "unknown", "style": "style1"},
  "h_1": {"color": "#5D2A1C", "texture": "unknown", "style": "style1"},
  "a": {"color": "#5D2A1C", "texture": "unknown", "style": "style1"},
  "c_1": {"color": "#5D2A1C", "texture": "unknown", "style": "style1"},
  "h_2": {"color": "#5D2A1C", "texture": "unknown", "style": "style1"},
  "ế": {"color": "#5D2A1C", "texture": "unknown", "style": "style1"}
}
\end{verbatim}
}
\end{mybox}
\caption{Prompt for Design Transfer.}
\label{fig:prompt_design_transfer}
\end{figure*}

\begin{figure*}[t]
\centering
\begin{mybox}{Prompt for Auto Cropping}
\textbf{Context}: I have a character segmentation model that detects individual character components (strokes, radicals, parts, etc.) but outputs them as separate bounding boxes rather than complete characters. I need help grouping these component detections into complete character bounding boxes.

\textbf{Task}: Group the individual component bounding boxes into clusters that represent complete characters.

\textbf{Input}:
\begin{itemize}
    \setlength\itemsep{0em}
    \item An image containing target text \& component-level bounding box detections (with numbering on each bounding box)
    \item Target text to detect (if not provided, run OCR on the image to get the target text)
\end{itemize}

\textbf{Expected Output}:
\begin{itemize}
    \setlength\itemsep{0em}
    \item Output a JSON dict, where keys have the exact same chars as in the detected text or provided text (except empty spaces), values are the indices of bounding boxes to merge
    \item For duplicate characters, append suffix starting with ``\_1''.
\end{itemize}

\textbf{Input \& Output Explanations:}
\begin{CJK*}{UTF8}{gbsn}
\begin{itemize}
    \setlength\itemsep{0em}
    \item Keys represent actual character text string (e.g., 你, 好, 世, 届, etc.) from the OCR or provided text
    \item Values are arrays of component indices that belong to that character from the image
    \item Component indices refer to the detection boxes in the provided image
\end{itemize}
\end{CJK*}

\textbf{Grouping Criteria:}
\begin{itemize}
    \setlength\itemsep{0em}
    \item Consider the spatial proximity of components for a complete character
    \item Ensure the merged final bounding boxes align with the target text sequence
    \item Follow the reading order (left-to-right, top-to-bottom, etc.) to group the component-level bounding boxes
    \item Do NOT use nested lists, e.g., [[7], [8], [9]]; use [7, 8, 9] instead
    \item Return JSON format only!
\end{itemize}

\begin{CJK*}{UTF8}{gbsn}
\textbf{CJK Example}
\begin{footnotesize}
\begin{verbatim}
Input text: "你好世界"
Output JSON:
{
  "你": [1, 4, 3, 2],
  "好": [3], 
  "世": [6, 7, 8],
  "届": [9],
}

\end{verbatim}
\end{footnotesize}
\end{CJK*}

\textbf{Latin Example}
\begin{verbatim}
Input text: "mississippi"
Output JSON:
{
  "m": [1],
  "i": [2, 3],         
  "s": [4],         
  "s_1": [5], 
  "i_1": [6,7], 
  "s_2": [8],
  ...
}
\end{verbatim}
\end{mybox}
\caption{Prompt for Auto Cropping.}
\label{fig:prompt_auto_cropping}
\end{figure*}

\end{document}